\documentclass{article}

\usepackage[table,xcdraw]{xcolor}
\usepackage{hyperref}
\usepackage{url}
\usepackage{xifthen}
\usepackage{makecell}
\usepackage{booktabs}
\usepackage{amsfonts}
\usepackage{nicefrac}
\usepackage{microtype}
\usepackage{amsmath}
\usepackage{multirow}
\usepackage{subcaption}
\usepackage{graphicx}
\usepackage[makeroom]{cancel}

\usepackage[accepted]{icml2024}

\def\N{\mathcal{N}}
\def\E{\mathbb{E}}
\def\R{\mathbb{R}}
\def\x{\mathbf{x}}
\def\hx{\hat{\x}}
\def\z{\mathbf{z}}
\def\w{\mathbf{w}}
\def\bw{\bar{\w}}
\def\a{\alpha}
\def\s{\sigma}
\def\ts{\tilde{\s}}
\def\veps{\varepsilon}
\def\vphi{\varphi}
\def\mnu{\mathbb{\nu}}
\def\D{\Delta}
\def\d{\partial}
\def\KL{{\mathrm{D}_\mathrm{KL}}}
\def\L{\mathcal{L}}
\def\Lrec{{\L_\mathrm{rec}}}
\def\Ldiff{{\L_\mathrm{diff}}}
\def\Lprior{{\L_\mathrm{prior}}}
\def\Lsimple{{\L_\mathrm{simple}}}

\newcommand{\TODO}[1][]{{\ifthenelse{\isempty{#1}}{\color{red}\bf TODO}{\color{red}\bf (TODO: #1)}}}

\icmltitlerunning{Neural Diffusion Models}

\begin{document}

\twocolumn[
\icmltitle{Neural Diffusion Models}



\icmlsetsymbol{equal}{*}

\begin{icmlauthorlist}
\icmlauthor{Grigory Bartosh}{uva}
\icmlauthor{Dmitry Vetrov}{ctr}
\icmlauthor{Christian A. Naesseth}{uva}
\end{icmlauthorlist}

\icmlaffiliation{uva}{University of Amsterdam}
\icmlaffiliation{ctr}{Constructor University, Bremen}

\icmlcorrespondingauthor{Grigory Bartosh}{g.bartosh@uva.nl}
\icmlcorrespondingauthor{Dmitry Vetrov}{dvetrov@constructor.university}
\icmlcorrespondingauthor{Christian A. Naesseth}{c.a.naesseth@uva.nl}

\icmlkeywords{diffusion, generative models, variational inference}

\vskip 0.3in
]



\printAffiliationsAndNotice{}  

\begin{abstract}
Diffusion models have shown remarkable performance on many generative tasks. Despite recent success, most diffusion models are restricted in that they only allow linear transformation of the data distribution. In contrast, broader family of transformations can help train generative distributions more efficiently, simplifying the reverse process and closing the gap between the true negative log-likelihood and the variational approximation. In this paper, we present Neural Diffusion Models (NDMs), a generalization of conventional diffusion models that enables defining and learning time-dependent non-linear transformations of data. We show how to optimise NDMs using a variational bound in a simulation-free setting. Moreover, we derive a time-continuous formulation of NDMs, which allows fast and reliable inference using off-the-shelf numerical ODE and SDE solvers. Finally, we demonstrate the utility of NDMs through experiments on many image generation benchmarks, including MNIST, CIFAR-10, downsampled versions of ImageNet and CelebA-HQ. NDMs outperform conventional diffusion models in terms of likelihood, achieving state-of-the-art results on ImageNet and CelebA-HQ, and produces high-quality samples.
\end{abstract}

\section{Introduction}
\label{sec:introduction}

Generative models are a powerful class of probabilistic machine learning models with a wide range of applications from e.g.\@ art and music to medicine and physics \citep{tomczak2022deep,creswell2018generative,papamakarios2021normalizing,yang2022diffusion}. Generative models learn to mimic the underlying probability distribution of a given data set and can generate novel samples that are similar to the original data. They can for example be used for data augmentation, as well as for unsupervised learning.

Diffusion models have emerged as a family of generative models that excel at several generative tasks \citep{sohl2015deep, ho2020denoising}. They parameterize the data model through an iterative refinement process, the \emph{reverse process}, that builds up the data step-by-step from pure noise. For training purposes an auxiliary noising process, the \emph{forward process}, is introduced that successively adds noise to data. The reverse process is then optimized to resemble the forward process. Despite success in various domains \citep{sohl2015deep, ho2020denoising,saharia2021image, popov2021grad, watson2022broadly,trippe2023diffusion}, a key limitation of most existing diffusion models is that they rely on a fixed and pre-specified forward process that is unable to adapt to the specific task or data at hand. At the same time there are many works \citep{hoogeboom2023blurring, rombach2022high, lipman2023flow} that improve performance of diffusion models by modifications of the forward processes.

In this paper we develop Neural Diffusion Models (NDMs), a general framework that enables non-linear, time-dependent and learnable data transformations. We extend the approach by \citet{song2021denoising} and construct the forward process as a non-Markovian sequence of latent variables; each latent variable is constructed through a transformation of the data to which we then inject noise. This is then leveraged in the corresponding reverse process. To train NDMs efficiently we generalize the diffusion objective while keeping it a simulation-free bound on the log-likelihood. Furthermore, we derive the time-continuous analogue of the objective function as well as the stochastic differential equation (SDE) and ordinary differential equation (ODE) corresponding to the reverse process.

We demonstrate how NDMs generalizes several existing diffusion models and then propose a new model with learnable transformations of data parameterized by a neural network. To illustrate the empirical properties of NDMs we provide experimental results on a synthetic data as well as on MNIST, CIFAR-10, downsampled ImageNet and CelebA-HQ image datasets. NDMs consistently outperforms baselines in terms of negative log-likelihood, achieving state-of-the-art results for diffusion models on ImageNet 32 and 64, as well as CelebA-HQ. 

The main motivation for NDMs is improved likelihood and density estimation, crucial for applications to compression \citep{mackay2003information}, semi-supervised learning \citep{dai2017good}, adversarial purification \citep{song2017pixeldefend}, and many others. However, for completeness we also study the impact of NDMs on image generation quality. We find that for small to medium number of steps NDMs achieves better image generation quality than denoising diffusion probabilistic models (DDPMs) \citep{ho2020denoising}, and comparable results for a large number of steps. 

Finally, we demonstrate that NDMs allows learning simpler generative dynamics like dynamical optimal transport, which conventional diffusion models are incapable of learning.

We summarize the contributions as follows:
\begin{enumerate}
    \item We propose neural diffusion models or NDMs, a new framework that generalizes conventional diffusion models in both discrete and continuous time settings.

    \item We develop an objective function to optimize NDMs that upper bounds the negative log-likelihood and study its properties.

    \item We demonstrate the utility of NDMs with learnable transformations in terms of consistently and significantly improved log-likelihood, as well as better or comparable generation quality. 
\end{enumerate}

\begin{table*}[!t]
\caption{Summary of existing diffusion models as instances of Neural Diffusion Models (NDM). See extended table in Appendix \ref{app:related}.}
\label{tab:generalization}
\centering
\begin{tabular}{lccl}
\toprule


Model 
& 
Distribution $q(\z_t|x)$ 
& 
NDM's $F(\x, t)$
& 
Comment
\\
\midrule


\makecell[l]{
    DDPM \citep{ho2020denoising} / \\
    DDIM \citep{song2021denoising}
}
& 
$\N\Big(\z_t; \a_t \x, \s^2_t I \Big)$ 
& 
$\x$
& 

\\
\\


\makecell[l]{
    Flow Matching OT \\
    \citep{lipman2023flow}
}
& 
$\N\Big(\z_t; \a_t \x, \s^2_t I \Big)$ 
& 
$\x$
& 
\makecell[l]{
    $\a_t = t$, \\
    $\s_t = 1 - (1 - \s_{\mathrm{min}})t$ 
}
\\
\\


\makecell[l]{
    VDM \\
    \citep{kingma2021variational}
}
& 
$\N\Big(\z_t; \a_t \x, \s^2_t I \Big)$ 
& 
$\x$
& 
\makecell[l]{
    $\a^2_t = \mathrm{sigmoid}(-\gamma_{\eta}(t))$, \\
    $\s^2_t = \mathrm{sigmoid}(\gamma_{\eta}(t))$ 
}
\\
\\


\makecell[l]{
    Soft Diffusion \\
    \citep{daras2022soft}
}
&
$\N\Big(\z_t; C_t \x, s_t^2 I \Big)$ 
& 
$C_t \x$
&
$\a_t = 1$, $\s^2_t = s^2_t$
\\
\\


\makecell[l]{
    LSGM \\
    \citep{vahdat2021score}
}
&
$\N\Big(\z_t; \a_t E(\x), \s_t^2 I \Big)$ 
&
$E(\x)$
& 
$p(x|z_0) = \N\Big( x; a D(z_0), \s^2 \Big)$
\\
\\


\bottomrule
\end{tabular}
\end{table*}

\section{Background}
\label{sec:background}

Diffusion models are generative models that make use of latent variables. Given a sample from the data distribution $\x \sim q(\x)$, we define a forward noising process that produces latent variables $\z_0, \z_1, \dots, \z_T$. In contrast, the reverse generative process reverts the forward process, starting by first generating the same latent variables and then data $\x$.

\begin{figure}[tp]
    \centering
    \parbox[b]{.49\textwidth}{
        \begin{subfigure}[b]{\linewidth}
            \includegraphics[width=\textwidth]{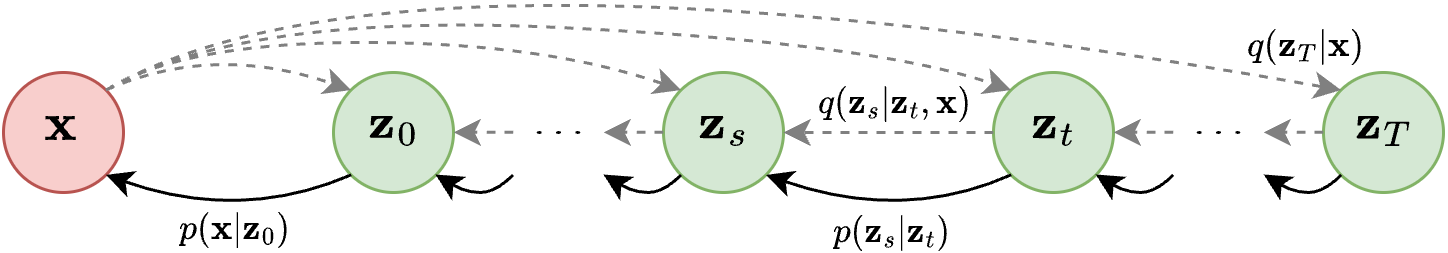}
            \caption{DDIM}
            \label{fig:ddim_graphical}
        \end{subfigure}
    }
    \hfill
    \parbox[b]{.49\textwidth}{
        \begin{subfigure}[b]{\linewidth}
            \includegraphics[width=\textwidth]{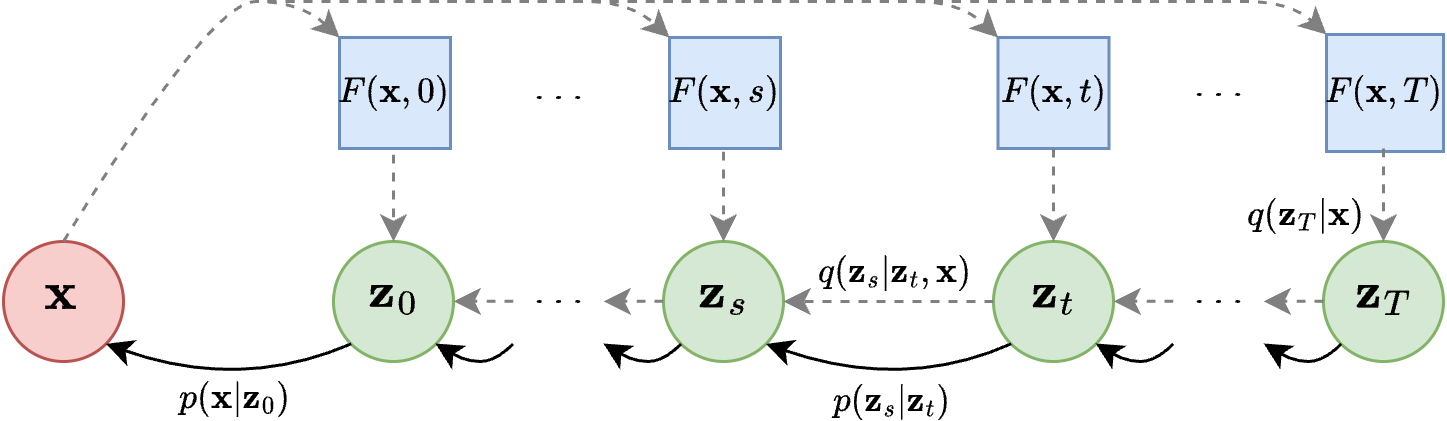}
            \caption{NDM}
            \label{fig:ndm_graphical}
        \end{subfigure}
    }
    \caption{The directed graphical models of DDIM and NDM.}
    \label{fig:graphical}
\end{figure}

The standard approach to specify the forward process is as a linear Gaussian Markov chain \citep{sohl2015deep, ho2020denoising}. However, we can also use an implicit definition of the forward process from \citet{song2021denoising}. This will turn out to be useful for our purposes and is what we focus on here. To construct the implicit forward process we first define the marginal distributions $q(\z_t|\x)$. Using these marginal distributions we can define the joint distribution of all latent variables $\z_0, \z_1, \dots, \z_T$ as follows:
\begin{align}
    q(\z_{0:T}|\x) &=  q(\z_T|\x) \prod_{t=1}^T q(\z_{t-1}|\z_t, \x), \nonumber\\
    \textrm{with} \quad &q(\z_{t-1}|\z_t, \x) \quad \textrm{such that} \nonumber \\ 
     q(\z_{t-1}|\x) &= \int q(\z_t|\x) q(\z_{t-1}|\z_t, \x) d\z_t.
    \label{eq:forward_process}
\end{align}
Here we make use of the posterior distribution $q(\z_{t-1}|\z_t, \x)$ instead of the regular forward distribution $q(\z_t|\z_{t-1})$. Due to the dependence also on the data $\x$ it is a non-Markovian forward process (see Figure~\ref{fig:ddim_graphical}). In general the forward process is considered fixed and has no trainable parameters. Moreover, it is specified in such a way that $q(\z_0|\x) \approx \delta(\z_0 - \x)$ and $q(\z_T|\x) \approx \N(\z_T; 0, I)$. So if $q(\z_{t-1}|\z_t)$ was available we could sample $\z_T \sim \N(\z_T; 0, I)$ and run the reverse process to get $\z_0 \sim q(\z_0) \approx q(\x)$. However, the distribution $q(\z_{t-1}|\z_t)$ depends implicitly on the data distribution $q(\x)$ and thus has a complicated form, so we instead approximate the reverse process using a Markov chain with distribution $p_{\theta}(\z_{0:T})$:
\begin{equation}
    p_{\theta}(\z_{0:T}) = p(\z_T) \prod_{t=1}^T p_{\theta}(\z_{t-1}|\z_t), 
    \label{eq:reverse_process}
\end{equation}
where $p(\z_T) = \N(\z_T; 0, I)$.

The combination of the forward process $q$ and the reverse process $p_\theta$ is a form of (hierarchical) variational autoencoder \citep{kingma2014auto, rezende2014stochastic}. Therefore, it can be trained by optimizing the usual variational bound on the negative log-likelihood. In the case of diffusion models, it can be written as follows (see Section A of \citet{ho2020denoising}):
\begin{align}
    \E_{q} &\Bigg[ \underbrace{\KL \Big( q(\z_T|\x) || p(\z_T) \Big)}_{\Lprior} - \underbrace{\log p_{\theta} (\x|\z_0)}_{\Lrec} \nonumber\\
    &+\underbrace{\sum_{t=1}^T \KL  \Big( q(\z_{t-1}|\z_t, \x) || p_{\theta} (\z_{t-1}|\z_t) \Big)}_{\Ldiff} \Bigg]
    \label{eq:ddim_elbo}
\end{align}

Since the process $q$ and the distribution $p_{\theta}(\z_T)=p(\z_T)$ are fixed, the prior term $\Lprior$ does not depend on the parameters $\theta$, so it can be omitted. The distribution $p_{\theta} (\x|\z_0)$ is often take to be a Gaussian distribution, with low variance, for continuous data and a dequantization distribution for discrete data. Thus, also the reconstruction term $\Lrec$ does not depend on the parameter $\theta$.

This means that the only part that depends on the model parameters $\theta$ is the diffusion term $\Ldiff$. It is a sum of Kullback–Leibler (KL) divergences between posterior distributions in the forward process $q(\z_{t-1}|\z_t, \x)$ and the distributions $p_{\theta} (\z_{t-1}|\z_t)$ from the reverse process. In the general case this KL divergence is intractable, so the standard choice here is to set the marginal conditional distributions to be Gaussian, i.e.\@ $q(\z_t|\x) = \N(\z_t; \a_t \x, \s_t^2 I)$. The posterior distribution then takes the form:
\begin{align}
     q(\z_s|\z_t, \x) &= \N \left( \z_s; \mu_{s|t}, \ts_{s|t}^2 I \right),
     \quad \textrm{for} \quad 0 \leq s \leq t \leq T, \nonumber\\
     \mu_{s|t} &= \a_s \x + \frac{\sqrt{\s_s^2 - \ts_{s|t}^2}}{\s_t} (\z_t - \a_t \x).
     \label{eq:ddim_posterior}
\end{align}
Note that here we allow for an arbitrary choice of time grid, i.e.\@ $s$ and $t$, whereas above it was equidistant. It is straightforward to check that such a posterior distribution satisfies (\ref{eq:forward_process}) for any $\ts_{s|t}^2 \leq \s_s^2$. The exact schedule of $\ts_{s|t}^2$ is a user design choice.

Finally, the reverse distribution is set to $p_{\theta} (\z_s|\z_t) = q(\z_s|\z_t, \hx_{\theta}(\z_t, t))$, where $\hx_{\theta}(\z_t, t)$ is the model's prediction of $\x$. Since $q(\z_s| \z_t, \x)$ and $p_{\theta}(\z_s|\z_t)$ are both Gaussian distributions, we can compute the KL divergences in $\Ldiff$ in closed form.

This choice of forward and reverse processes, resulting in analytic expressions for the diffusion terms given data, is what makes diffusion models a \emph{simulation-free} approach. Simulation-free means that we do not have to sample all latent variables for each optimization step. Rather than calculating all individual terms in $\Ldiff$, we can uniformly sample $t$ and optimize only a subset of KL divergences using stochastic gradient descent.

By choosing a specific value for $\ts_{s|t}^2$, we can obtain equality between the processes of DDPM and DDIM (see section 4.1 of \citet{song2021denoising}). Furthermore, as \citet{song2021scorebased} demonstrated, when the number of steps $T$ in DDPM goes to infinity, we can transition to continuous time. In this scenario, the reverse process can be described using a Stochastic Differential Equation (SDE):
\begin{align}
    d \z_t &= [r(t) \z_t - g^2(t) s_\theta (\z_t, t)] dt + g(t) d \w_t, \nonumber \\ & s_\theta (\z_t, t) = \frac{\a_t \hx_{\theta}(\z_t, t) - \z_t}{\s_t^2}, \quad r(t) = \frac{d \log \a_t}{dt}, \nonumber \\
    &g^2(t) = \frac{d \s_t^2}{dt} - 2 \frac{d \log \a_t}{dt} \s_t^2,
    \label{eq:ddpm_reverse_sde}
\end{align}
with time running backwards from $t=1$ to $t=0$. This formulation allows us to switch to the equivalent ODE and to use different SDE and ODE solvers for sampling and density estimation.

\begin{figure*}[!t]
\begin{minipage}{0.49\textwidth}
\begin{algorithm}[H]
\caption{Learning NDM}
\label{alg:training}
\begin{algorithmic}
    \REQUIRE $q(\x)$, $F_\vphi$, $\hx_\theta$
    \FOR{learning iterations}
        \STATE $\x \sim q(\x)$, $t \sim U[1, T]$, $\veps \sim \N(0, I)$
        \STATE $\z_t \sim q_\vphi(\z_t|\x)$
        \STATE $\L = \Lrec + \Ldiff + \Lprior$
        \STATE Gradient step on $\theta$ and $\vphi$ w.r.t. $\L$
    \ENDFOR
\end{algorithmic}
\end{algorithm}
\end{minipage}
\hfill
\begin{minipage}{0.49\textwidth}
\begin{algorithm}[H]
\caption{Sampling from NDM}
\label{alg:sampling}
\begin{algorithmic}
    \REQUIRE $F_\vphi$, $\hx_\theta$
    \STATE $\z_T \sim \N(0, I)$
    \FOR{$t = T, \dots, 1$}
        \STATE $\hx = \hx_\theta(\z_t, t)$
        \STATE $\z_{t-1} \sim q_\vphi(\z_{t-1}|\z_t, \hx)$
    \ENDFOR
    \STATE $x \sim p(\x|\z_0)$
\end{algorithmic}
\end{algorithm}
\end{minipage}
\end{figure*}

\section{Neural diffusion models}
\label{sec:model}

In theory, we can view diffusion models as a special type of hierarchical variational autoencoders (VAE). From this perspective, the conventional diffusion model resembles a VAE with a fixed variational distribution, in which the latent variables are inferred using scaling of data points and injecting of Gaussian noise. Such a formulation limits diffusion models in terms of the flexibility of the latent space. Introducing a more flexible (and learnable) distribution of latent variables could effectively reduce the gap between the log-likelihood and the variational bound. In practical terms, a more flexible forward process might simplify the task of learning the reverse (generative) process, thereby enhancing model quality. To overcome this limitation of conventional diffusion models, we propose a general form of transformations of data that allows to define and learn distributions on the latent space.

In this section, we introduce the Neural Diffusion Models (NDMs) -- a simulation-free framework that generalises conventional diffusion models. The key idea in NDMs is to apply a time-dependent transformation $F_\vphi(\x, t)$ to the data $\x$ at each step of forward process before injecting noise. Previous diffusion models arise as special cases when the data transformation is either linear, time-independent, or pre-defined non-linear (see Table~\ref{tab:generalization}). In contrast, the NDM can work with any time-dependent transformation of data and may be learned end-to-end. In Section~\ref{sec:experiments} we provide experimental results with $F_\vphi(\x, t)$ parameterized by neural networks.

\subsection{Model definition and variational objective}

We introduce NDMs constructively. First, we define the desired marginal distributions:
\begin{align}
     q_\vphi(\z_t|\x) = \N \Big( \z_t; \a_t F_\vphi(\x, t), \s_t^2 I \Big),
     \label{eq:ndm_marginal}
\end{align}
where $F_\vphi(\x, t): \R^d \times [0, T] \mapsto \R^d$ is a function parameterized by $\vphi$ that applies a time-dependent transform to the data point $\x$. We adapt the approach from DDIM, as described in Section~\ref{sec:background}, and choose the following posterior distribution that satisfies (\ref{eq:ndm_marginal}) (we provide derivation and proof in Appendix~\ref{app:posterior}):
\begin{align}
     &q_\vphi(\z_s|\z_t, \x) = \N \left( \z_s; \mu^{F_\vphi}_{s|t}, \ts_{s|t}^2 I \right), 
     \label{eq:ndm_posterior}\\
     \mu^{F_\vphi}_{s|t} &= \a_s F_\vphi(\x, s) + \frac{\sqrt{\s_s^2 - \ts_{s|t}^2}}{\s_t} \Big( \z_t - \a_t F_\vphi(\x, t) \Big),\nonumber
\end{align}
for $0 \leq s \leq t \leq T$ where $\ts_{s|t}^2 \leq \s_s^2$ is a design choice. Using this posterior we can define an implicit forward process according to (\ref{eq:forward_process}) (see Figure~\ref{fig:ndm_graphical}). This forward process provides access to both marginal and posterior distributions just like in the DDIM framework \citep{song2021denoising}. The corresponding NDM variational objective  has the following form:
\begin{align}
    \E_{q_\vphi} &\Bigg[ \underbrace{\KL \Big( q_\vphi(\z_T|\x) || p(\z_T) \Big)}_{\Lprior}- \underbrace{\log p_{\theta} (\x|\z_0)}_{\Lrec} \nonumber\\
    &+
    \underbrace{\sum_{t=1}^T \KL  \Big( q_\vphi(\z_{t-1}|\z_t, \x) || p_{\theta} (\z_{t-1}|\z_t) \Big)}_{\Ldiff}  \Bigg].
    \label{eq:elbo}
\end{align}

While the objective has the same form as in DDIM (\ref{eq:ddim_elbo}), the individual terms are different. If the transformation $F_\vphi(\x, t)$ is actually parameterized by learnable parameters $\vphi$, the prior term $\Lprior$ and the reconstruction term $\Lrec$ depend on the parameter $\vphi$ as well. Therefore, in that case these terms cannot be excluded from the optimization process. 

For the standard parameterization of the reverse process through approximate posteriors $p_{\theta} (\z_s|\z_t) = q_\vphi(\z_s|\z_t, \hx_{\theta}(\z_t, t))$ the KL divergences in the diffusion term $\Ldiff$ are (see Appendix~\ref{app:objective}):
\begin{align}
    &\KL \Big( q_\vphi(\z_s|\z_t, \x) || p_{\theta} (\z_s|\z_t) \Big) = 
    \nonumber \\ 
    &\frac{1}{2 \ts_{s|t}^2} \Bigg\| 
     \a_s \Big( F_\vphi(\x, s) - F_\vphi(\hx_{\theta}(\z_t, t), s) \Big) +\nonumber \\
    & ~
     \frac{\sqrt{\s_s^2 - \ts_{s|t}^2}}{\s_t} \a_t \Big( F_\vphi(\hx_{\theta}(\z_t, t), t) - F_\vphi(\x, t) \Big) \Bigg\|_2^2.
     \label{eq:ndm_kl}
\end{align}

Note a distinction between the objectives of NDM and DDIM here. In the case of DDIM, the model tries to accurately predict the data point $\x$. In contrast, NDM aims to predict the \textit{transformed} data point $F_\vphi(\x, t)$. Despite this change, NDM's optimization remains simulation-free, so we can efficiently train the NDM by sampling time steps and calculating corresponding KL divergences. We summarise the training and sampling procedures in Algorithms \ref{alg:training} and \ref{alg:sampling}.

Given that NDM is a generalization of DDIM, we can leverage the same techniques for inference. Specifically, we can adjust the number of intermediate time steps, the schedule of $\ts_{s|t}^2$ as well as sampling with various dynamics, including a deterministic dynamic corresponding to $\ts_{s|t}^2=0$.

\subsection{Continuous time NDMs}
\label{sec:cont_ndm}

We previously formulated NDMs in the discrete time setting with $T$ steps. However, like conventional diffusion models, we can let the number of steps $T$ go to infinity and switch to continuous time. In this case, the set of time steps $\{0, 1, \dots, T\}$ transforms to the range $[0, 1]$ and the diffusion term of the objective reduces to an expectation over time (see derivation in Appendix~\ref{app:cont_objective}):
\begin{align}
    &\Ldiff = 
     \nonumber \\
    & \E_{q(\x)u(t)q(\z_t|\x)}\Bigg[\frac{1}{g^2(t)} \Bigg\| 
    \a_t \Big( \dot{F}_\vphi (\x, t) - \dot{F}_\vphi \big(\hx_{\theta}(\z_t, t), t\big) \Big) \nonumber \\
    & 
    +\frac{1}{2} \Bigg( \frac{\d \s_t^2}{\d t} -  2 r(t) \s_t^2 + g^2(t) \Bigg) \cdot \nonumber \\
    &\quad \quad \quad \cdot \Big( s(\x, \z_t, t) - s \big(\hx_{\theta}(\z_t, t), \z_t, t \big) \Big)
    \Bigg\|_2^2 ~\Bigg], 
    \label{eq:ldiff_cont}
\end{align}
where $r(t) = \frac{\d \log \a_t}{\d t}$, $g^2(t) = \dot{\mnu}_t \s_t^2$, and ${s(\x, \z_t, t) = \frac{\a_t F_\vphi(\x, t) - \z_t}{\s_t^2}}$.

Similar to training a discrete time NDM, we can train a continuous time NDMs by sampling time. In our experiments we use importance sampling \citep{song2021maximum} and sample time from a distribution proportional to $\frac{1}{g^2(t)}$.

Note, that we may not have access to the partial derivative of the transformation $F_\vphi(\cdot, t)$ with respect to $t$ in closed form. However, for any differentiable $F_\vphi(\cdot, t)$ we can use Jacobian-Vector product \citep{smale1974differential} to obtain this derivative.

The discrete time reverse process also becomes a continuous time process, described by a Stochastic Differential Equation (SDE). If we parameterize the noise injection in the posterior distribution as $\ts_{s|t}^2 = \sigma_s^2 (1 - e^{\mnu_s - \mnu_t})$, we obtain the following SDE (see derivation in Appendix~\ref{app:sde}):
\begin{align}
    d \z_t = &\Big[ \a_t \dot{F}_\vphi(\hx_{\theta}(\z_t, t), t) + r(t) \z_t \nonumber \\
    - &\frac{1}{2} \Big( g^2(t) - 2 r(t) \s_t^2 \Big) s_\theta(\z_t, t)  \Big] d t + g(t) d \w, 
    \label{eq:sde}
\end{align}
where $s_\theta (\z_t, t) = \frac{\a_t F_\vphi(\hx_{\theta}(\z_t, t), t) - \z_t}{\s_t^2}$.

By changing the function $\mnu_t$, we can obtain different dynamics. In the extreme case where $\mnu_t$ is equal to a constant we have deterministic dynamics described by an ODE. This enables the use of SDE or ODE solvers for inference. Moreover, we can estimate densities by considering the model as a continuous normalizing flow \citep{chen2018neural} in the deterministic case.

\section{Experiments}
\label{sec:experiments}

\begin{figure*}[tp]
    \centering
    \parbox{.22\textwidth}{
        \begin{subfigure}[b]{\linewidth}
            \includegraphics[width=\textwidth]{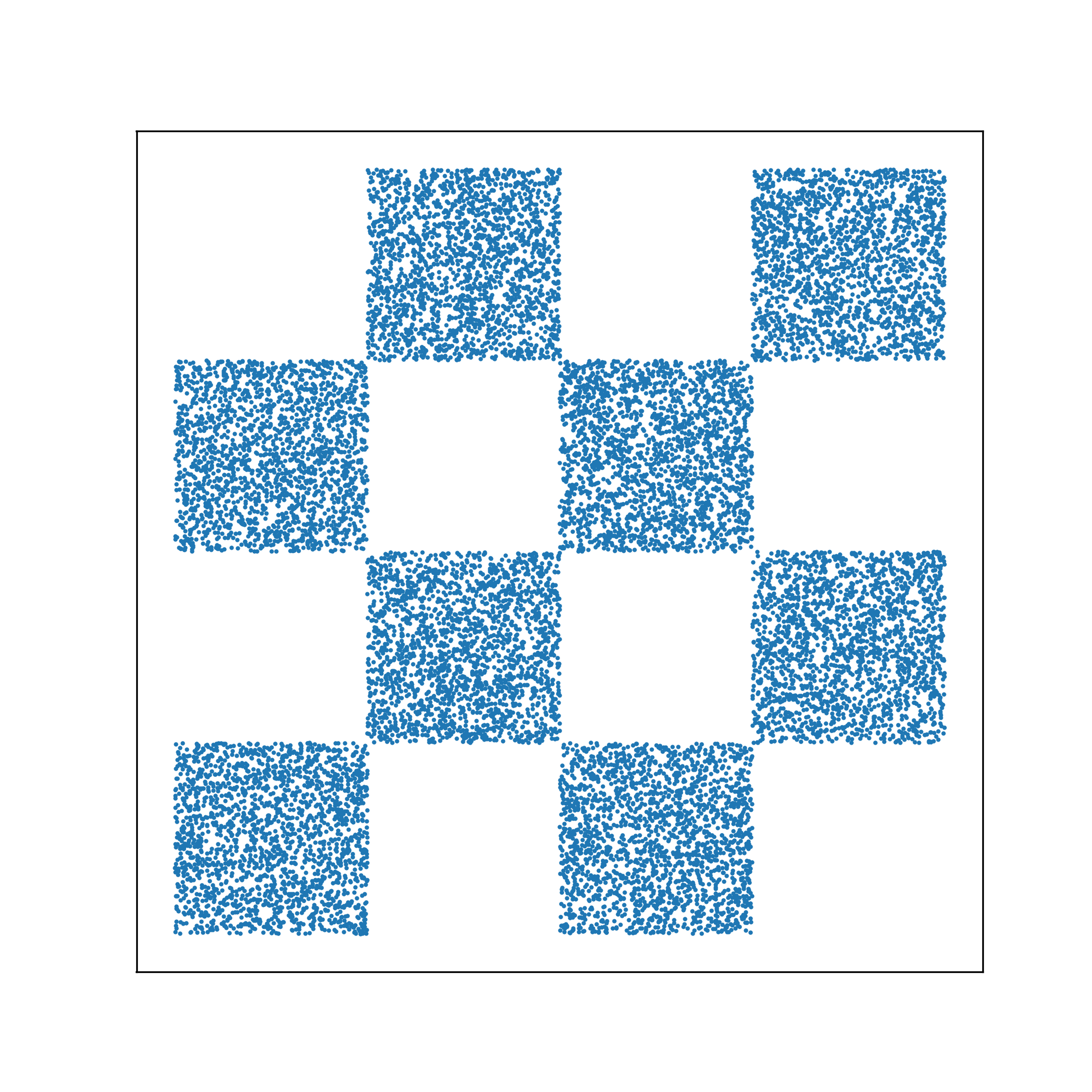}
        \end{subfigure}\\
        \begin{subfigure}[b]{\linewidth}
            \includegraphics[width=\textwidth]{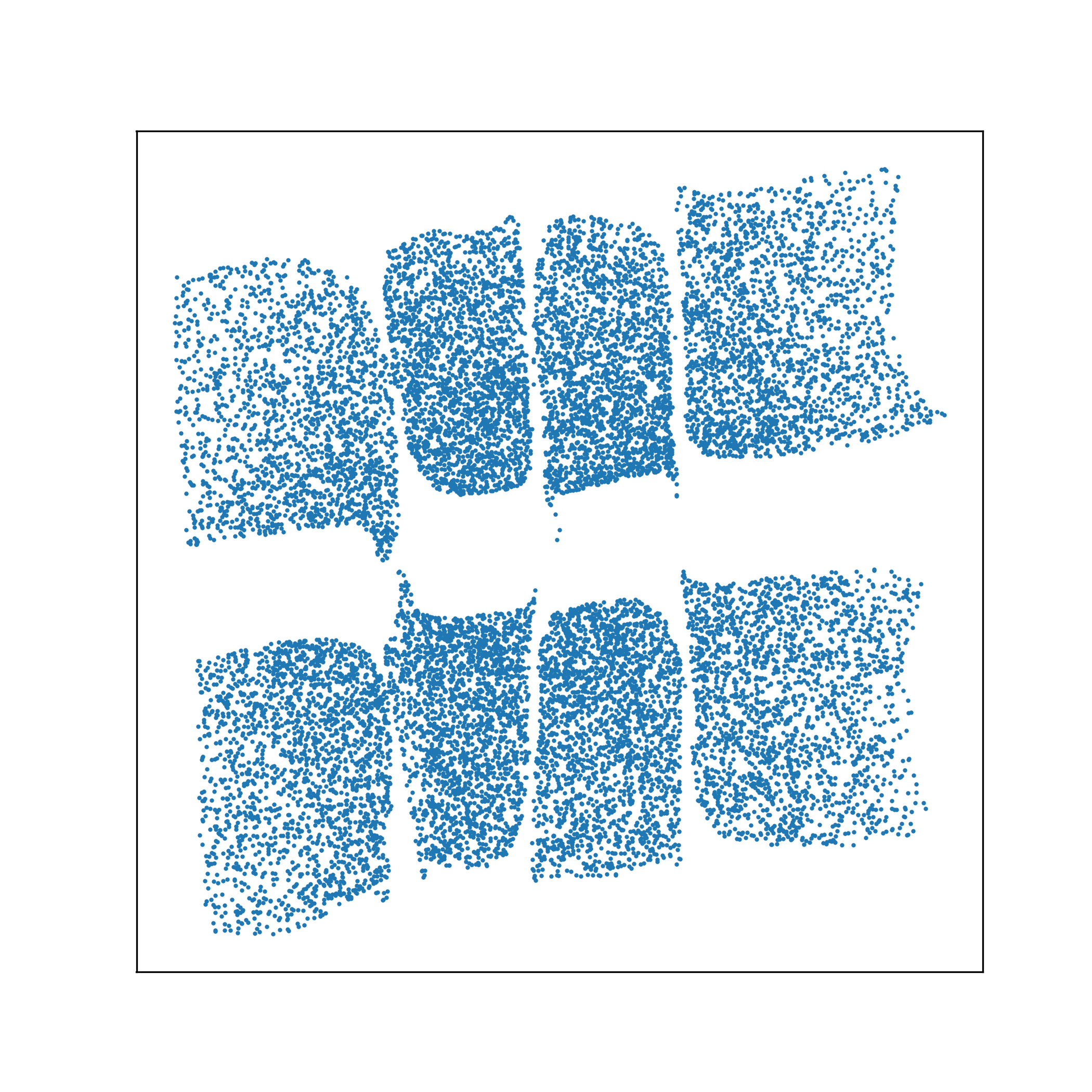}
            \caption{\emph{Top:} Data $\x$. \emph{Bottom:} $F_\vphi(\x,T)$, transformed data.}\label{fig:transforms_checkerboard}
        \end{subfigure}
    }
    \hfill
    \parbox{.77\textwidth}{
        \begin{subfigure}{\linewidth}
            \includegraphics[width=\textwidth]{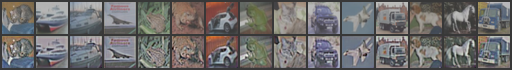}
            \caption{\emph{Top:} CIFAR data samples. \emph{Bottom:} $F_\vphi(\x,T)$, transformed data samples.}\label{fig:transforms_cifar}
        \end{subfigure}\\
        \begin{subfigure}{\linewidth}
            \includegraphics[width=\textwidth]{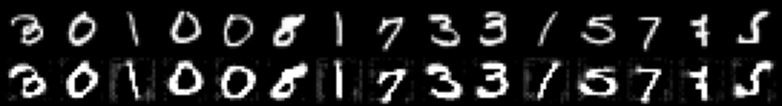}
            \caption{\emph{Top:} MNIST data samples. \emph{Bottom:} $F_\vphi(\x,T)$, transformed data samples.}\label{fig:transforms_mnist}
        \end{subfigure}\\
        \begin{subfigure}{\linewidth}
            \includegraphics[width=\textwidth]{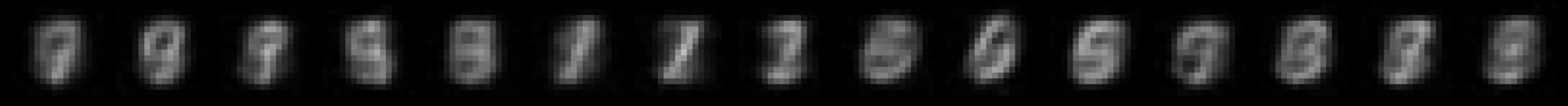}
            \caption{DDPM predictions $\hx_{\theta}(\z_T, T)$ for $\z_T \sim \N(\z_T; 0, I)$.}\label{fig:ddpm_predictions}
        \end{subfigure}\\
        \begin{subfigure}{\linewidth}
            \includegraphics[width=\textwidth]{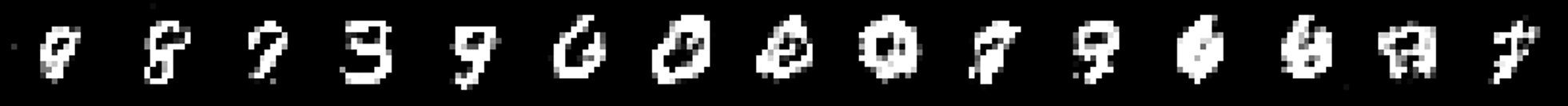}
            \caption{NDM predictions $\hx_{\theta}(\z_T, T)$ for $\z_T \sim \N(\z_T; 0, I)$.}\label{fig:ndm_predictions}
        \end{subfigure}
    }
    \caption{Learned transforms for the 2D checkerboard distribution (\emph{left}). Learned transforms for CIFAR-10 and MNIST (\emph{top right}), as well as predictions for MNIST (\emph{bottom right}). NDM learns useful forward transformations and more accurately predicts the data from injected noise.}\label{fig:transforms_predictions}
\end{figure*}

\begin{table*}[!t]
\caption{We report Negative Log-Likelihood (NLL) in Bits Per Dimension (BPD) on the test sets for CIFAR10, ImageNet 32x32, and ImageNet 64x64. Our results were obtained using the continuous-time formulation of our model, integrated via the corresponding Ordinary Differential Equation (ODE), as detailed in Section~\ref{sec:cont_ndm}.}
\label{tab:nll}
\centering
\begin{tabular}{llccc}
\toprule
Model &  & \textbf{CIFAR10} & \textbf{ImageNet 32} & \textbf{ImageNet 64}  \\ \cmidrule(r){1-1} \cmidrule(r){3-5}
DDPM \citep{ho2020denoising} &  & $3.69$ &  & \\
Improved DDPM \citep{nichol2021improved} &  & $2.94$ &  & $3.54$ \\
VDM \citep{kingma2021variational} &  & $\mathbf{2.65}$ & $3.72$ & $3.40$ \\
Score SDE \citep{song2021scorebased} &  & $2.99$ &  & \\
Score Flow \citep{song2021maximum} &  & $2.83$ & $3.76$ & \\
NDM (ours) &  & $2.70$ & $\mathbf{3.55}$ & $\mathbf{3.35}$ \\
\bottomrule
\end{tabular}
\end{table*}

We present empirical results for the proposed Neural Diffusion Models with learnable transformations on a synthetic datasets as well as multiple image datasets. Qualitatively, NDMs learn transformations that simplify the data distribution, leading to predictions of $\x$ that are more aligned with the data. Quantitatively, NDMs consistently outperform the baseline in terms of likelihood, achieving state-of-the-art diffusion model results for ImageNet and CelebA-HQ. Moreover, for a small to medium number of steps, NDMs achieve better image generation quality than DDPM, while being comparable for a large number of steps. We also provide a proof of concept experiment that demonstrates that NDMs can learn simple generative trajectories, something conventional diffusion models are incapable of learning.

\subsection{Implementation details}

We demonstrate NDMs with learnable transformations on the MNIST \citep{deng2012mnist}, CIFAR-10 \citep{krizhevsky2009learning}, downsampled ImageNet \citep{deng2009imagenet, van2016pixel} and CelebA-HQ-256 \citep{karras2017progressive} datasets. In all experiments we use same neural network architectures to parameterize both the generative process and the transformations $F_\vphi$. In experiments with images we use the U-Net architecture from \citet{dhariwal2021diffusion}. To ensure consistency with \citet{song2021scorebased, song2021maximum}, we apply horizontal flipping as a data augmentation technique for training models on CIFAR-10 and ImageNet. Unless otherwise stated, we utilize the DDPM variance-preserving schedule of noise injection for $\a_t$ and $\s^2_t$. For density estimation of discrete data we use uniform dequantization.

In the experiments we report negative log-likelihood (NLL) in bits per dimension (BPD), negative evidence lower bound (NELBO) (\ref{eq:elbo}), and sample quality as measured by the Frechet Inception Distance (FID) \citep{heusel2017gans}. We calculate NLL by integrating the corresponding ODEs using the RK45 solver from \citet{dormand1980family}, and both NLL and NELBO are calculated on test data. For FID we report the score computed using 50k generated images.

In Section \ref{sec:model} we parameterize the reverse process through $\hx_\theta$ function. However, in practice we reparameterize the generative process in terms of prediction of injected noise. For a detailed description of parameterizations and other experimental details, please refer to Appendix~\ref{app:implementation}.

\subsection{Learned transformations}

Let us examine some of the transformations that NDM learns. Figure~\ref{fig:transforms_checkerboard}-\ref{fig:transforms_mnist} illustrates the transformations that NDM learns for the 2D checkerboard distribution, MNIST, and CIFAR-10 datasets. For the checkerboard, we observe that $F_\vphi$ learns to transform the interleaved pattern into a non-interleaved one. In the case of the grayscale digits of the MNIST dataset, $F_\vphi$ learns to highlight the distinctive features of the numbers. It thickens the lines and even creates bubbles at the corners. For the color images of CIFAR-10, $F_\vphi$ learns to increase the image contrast. In all cases, our model learns a way to simplify the data distribution. These transformations may enable the reverse process to transition more smoothly from simple distributions to complex ones.

Furthermore, we would like to emphasize the difference between the predictions of $\x$ that NDM and DDPM makes. Figure~\ref{fig:ddpm_predictions} and Figure~\ref{fig:ndm_predictions} shows the predictions $\hx_{\theta}(\z_T, T)$ generated by NDM and DDPM models trained on the MNIST dataset. In each case, the model samples from a standard normal distribution $\z_T \sim \N(\z_T; 0, I)$ and based on this value tries to predict $\x$. Therefore, we do not expect these samples to be of high quality. However, as we can see, NDM's predictions of $\x$ are much more similar to the data distribution than DDPM's predictions. We attribute this behavior to the fact that our model aims to predict not the datapoint $\x$, but the transformed datapoint $F_\vphi(\x, t)$. Thus, to make better predictions of the transformed datapoint, it may be critical to generate predictions of $\x$ that resemble real data. Any deviation from the $\x$-distribution is exaggerated by the transformation and thus less likely to happen for NDM's predictions.

In Appendix \ref{app:additional} we provide additional samples for terminal and intermediate timesteps.

\begin{table*}[!t]
\caption{Performance comparison of the DDPM and NDM on CIFAR-10 and ImageNet 32 datasets. We report FID scores for DDPM-style (FID) and DDIM-style ($\textrm{FID}^*$) sampling procedures.}
\label{tab:results}
\centering
\begin{tabular}{lllrrrrrrrrr}
\toprule
                                         &                             &                          & \multicolumn{4}{c}{\textbf{CIFAR-10}}                                                                                & \multicolumn{1}{c}{}     & \multicolumn{4}{c}{\textbf{ImageNet 32}}                                                                             \\ \cmidrule(r){4-7} \cmidrule(r){9-12}
Steps                                    & Model                       &                          & \multicolumn{1}{c}{NLL $\downarrow$} & \multicolumn{1}{c}{NELBO $\downarrow$} & \multicolumn{1}{c}{FID $\downarrow$} & \multicolumn{1}{c}{$\textrm{FID}^*$ $\downarrow$} & \multicolumn{1}{c}{}     & \multicolumn{1}{c}{NLL $\downarrow$} & \multicolumn{1}{c}{NELBO $\downarrow$} & \multicolumn{1}{c}{FID $\downarrow$} & \multicolumn{1}{c}{$\textrm{FID}^*$ $\downarrow$} \\ \cmidrule(r){1-2} \cmidrule(r){4-7} \cmidrule(r){9-12}
                                         & DDPM                        &                          & 3.11                                 & 3.18                                   & \textbf{11.44}                                 & \textbf{13.35}                         &                          & 3.89                                 & 3.95                                   & \textbf{16.18} & \textbf{19.08}                                 \\
\multirow{-2}{*}{1000}                   & \cellcolor[HTML]{EFEFEF}NDM & \cellcolor[HTML]{EFEFEF} & \cellcolor[HTML]{EFEFEF}\textbf{3.02}         & \cellcolor[HTML]{EFEFEF}\textbf{3.03}           & \cellcolor[HTML]{EFEFEF}11.82         & \cellcolor[HTML]{EFEFEF}13.79 & \cellcolor[HTML]{EFEFEF} & \cellcolor[HTML]{EFEFEF}\textbf{3.79}         & \cellcolor[HTML]{EFEFEF}\textbf{3.82}           & \cellcolor[HTML]{EFEFEF}17.02 & \cellcolor[HTML]{EFEFEF}19.76         \\ \cmidrule(r){1-2} \cmidrule(r){4-7} \cmidrule(r){9-12}
                                         & DDPM                        &                          & 5.02                                 & 5.13                                   & 37.83                                 & \textbf{19.89}                         &                          & 6.28                                 & 6.42                                   & 53.51 & \textbf{26.47}                                 \\
\multirow{-2}{*}{10}                     & \cellcolor[HTML]{EFEFEF}NDM & \cellcolor[HTML]{EFEFEF} & \cellcolor[HTML]{EFEFEF}\textbf{4.63}         & \cellcolor[HTML]{EFEFEF}\textbf{4.74}           & \cellcolor[HTML]{EFEFEF}\textbf{31.56}         & \cellcolor[HTML]{EFEFEF}22.20 & \cellcolor[HTML]{EFEFEF} & \cellcolor[HTML]{EFEFEF}\textbf{5.81}         & \cellcolor[HTML]{EFEFEF}\textbf{5.94}           & \cellcolor[HTML]{EFEFEF}\textbf{45.38} & \cellcolor[HTML]{EFEFEF}29.95         \\ \midrule
                                         & DDPM                        &                          & 8.78                                 & 8.98                                   & \textbf{43.85}                                 & 17.73                         &                          & 10.99                                 & 11.23                                   & \textbf{58.35} & 25.53                                 \\
\multirow{-2}{*}{1000 $\rightarrow$ 10}  & \cellcolor[HTML]{EFEFEF}NDM & \cellcolor[HTML]{EFEFEF} & \cellcolor[HTML]{EFEFEF}\textbf{8.58}         & \cellcolor[HTML]{EFEFEF}\textbf{8.81}           & \cellcolor[HTML]{EFEFEF}48.41         & \cellcolor[HTML]{EFEFEF}\textbf{16.96} & \cellcolor[HTML]{EFEFEF} & \cellcolor[HTML]{EFEFEF}\textbf{10.78}         & \cellcolor[HTML]{EFEFEF}\textbf{11.06}           & \cellcolor[HTML]{EFEFEF}62.12 & \cellcolor[HTML]{EFEFEF}\textbf{23.77}        \\
\bottomrule
\end{tabular}
\end{table*}

\subsection{Image generation}

Next, we evaluate NDMs with learnable transformations quantitatively. We train continuous time NDM on MNIST, CIFAR-10, and downsampled ImageNet datasets. Table \ref{tab:nll} summarizes our results, reporting NLL. NDMs demonstrates performance on CIFAR-10 that is comparable with the baselines and outperforms baselines on ImageNet.

Then, we compare NDM with the DDPM baseline on MNIST, CIFAR-10, and ImageNet 32 datasets. To ensure a fair comparison, when implementing DDPM we use an NDM with fixed identity transformation $F_\vphi(\x,t)=\x$. Therefore, we train both models with the same objective (\ref{eq:elbo}) and hyperparameters. The first part of Table \ref{tab:results} summarizes our results, reporting NLL, NELBO (\ref{eq:elbo}), and FID score. NDM demonstrates comparable sample quality with the baseline on all datasets and consistently outperforms the baseline on NLL and NELBO, especially for smaller numbers of steps. This improvement may be attributed to NDM's ability to fit distributions of the forward process and simplify the denoising task for the reverse process.

We also compare NDM with DDPM in a setup where we train both models with $T=1000$ steps and then sample with fewer steps. The second part of Table \ref{tab:results} summarizes our results, which are consistent with the corresponding numbers of steps used during training. However, in absolute values, both models show worse performance when we decrease the number of steps, and NDM demonstrates a more severe degradation. This observation is especially noticeable for small numbers of steps, such as $T=10$, where NDM has a better FID score than DDPM when trained with $10$ steps, but a worse FID score when the number of steps is decreased from $1000$ to $10$. From this, we conclude that although NDM can in principle work with reduced number of steps it is less robust to such modifications compared to DDPM.

Finally, we demonstrate that NDMs may be successfully combined with LSGM \citep{vahdat2021score}. For this experiment we replaced the linear diffusion in the LSGM baseline for CelebA-HQ-256 with NDMs featuring the learnable $F_\vphi$. We parameterise $F_\vphi$ with the same neural network architecture as baseline's architecture for parameterisation of diffusion. Table \ref{tab:lsgm} demonstrates that NDMs have better likelihood estimation and sample quality.


In Appendix \ref{app:dot} we provide a proof of concept experiment, which demonstrates that we can learn simpler generative dynamics compared to conventional diffusion models. In this experiment we restrict the reverse process to learn dynamic optimal transport trajectories. It is not possible to match such a reverse process with a predefined forward process, but NDM allows to capture the data distribution with the simpler generative dynamics.

See Appendix \ref{app:additional} for additional results and ablation studies.

\begin{table}[!t]
\caption{Generative results on CelebA-HQ-256 for LSGM and NDM with learnable transformations in the latent space of VAE.}
\label{tab:lsgm}
\centering
\begin{tabular}{llrr}
\toprule
Model &  & NLL $\downarrow$ & FID $\downarrow$  \\ \cmidrule(r){1-1} \cmidrule(r){3-4}
LSGM \citep{vahdat2021score} &  & $\leq 0.70$ & $7.22$ \\
Latent NDM (ours) &  & $\mathbf{\leq 0.65}$ & $\mathbf{7.18}$ \\
\bottomrule
\end{tabular}
\end{table}

\section{Related work}
\label{sec:related_work}

NDMs build on diffusion probabilistic models originally proposed by \citet{sohl2015deep}, which can be considered as an instance of (hierarchical) variational autoencoders (VAEs) \citep{kingma2014auto, rezende2014stochastic}. Recently, the theory of diffusion models was extended to deterministic sampling \citep{song2021denoising} and continuous time \citep{song2021scorebased}. These results allowed to reach impressive performance in image generation tasks \citep{ho2020denoising, song2021scorebased, dhariwal2021diffusion, kingma2021variational}. However, most existing diffusion models have a significant limitation in that they rely on a pre-specified and simple noise injection process that is unable to adapt to the specific task or data at hand. To overcome this, researchers have explored ways to generalize diffusion models.

Various papers have since proposed ways to speed up sampling from diffusion models. \citet{tachibana2021taylor} and \citet{liu2022pseudo} proposed alternative SDE and ODE solvers. \citet{xiao2021tackling} proposed replacing simple Gaussian distributions at each generation step with distributions learned by GANs \citep{goodfellow2014generative}. Some works proposed methods like iterative distillation with a reduction in the number of steps \citep{salimans2022progressive} and iterative straightening of trajectories \citep{liu2023flow, liu2022rectified}. While these methods change the generative process, they are compatible with NDMs.

Several papers proposed constructing the process of data corruption not by noise injection, but rather by blurring \citep{rissanen2023generative, daras2022soft, hoogeboom2023blurring} or through another linear transformation \citep{singhal2023where}. Another line of work modifies directly the dynamics of diffusion models through mapping the data into the latent space of VAE \citep{vahdat2021score, rombach2022high}, hierarchical VAE \citep{gu2023fdm} or normalizing flow \citep{kim2022maximum} models and then runs standard linear diffusion. As we demonstrate in Tables \ref{tab:generalization}, these arise as distinct special cases of NDMs for specific choices of the transformation $F_\vphi$.

In another line of works \citep{de2021diffusion, wang2021deep, peluchettinon}, finite-time diffusion constructions were proposed using diffusion bridge theory to address the approximation error incurred by infinite-time denoising constructions. While such approaches allow learning forward transformations, they require inferring all latent variables for each optimization step. This limitation break the simulation-free paradigm and can make these models expensive to train. NDM in contrast allows learning forward transformations efficiently and simulation-free.

Inspired by diffusion models, several works \citep{lipman2023flow, neklyudov2022action} have proposed simulation-free objectives for training continuous normalizing flows. These approaches are similar to diffusion models as they rely on the idea of reversing a predefined corruption process. Later, some works \citep{albergo2023building, lee2023minimizing} extended these ideas and proposed to learn the forward process. However, although NDMs and these works are similar in spirit, they differ in that they optimize the forward process specifically to obtain straight generative trajectories, while in our approach we optimize learnable forward process to minimize variational bound on NLL, which not necessarily leads to straight generative trajectories.

In concurrent and independent work, \citet{nielsen2024diffenc} introduced DiffEnc, also adding time-dependent transformations in diffusion models. While the underlying idea is similar, there are some differences between DiffEnc and NDM. The two approaches utilize different parameterisations and noise injection schedules. In addition, DiffEnc approximates time derivatives of the data transformations leading to biased stochastic gradients, while NDM calculates exact time derivatives using JVP.

In Appendix \ref{app:related}, we provide further discussion, details and comparisons with related works.

\section{Limitations}
\label{sec:limitations}

Compared to conventional diffusion models, NDMs with learnable transformations have twice as many parameters, which results in slower training. Specifically, in experiments on images, NDMs with learnable transformations take approximately $2.3$ times longer than DDPM to train. However, no additional techniques where necessary to ensure stable training of NDMs. Additionally, in Appendix \ref{app:additional}, we provide an ablation study demonstrating that performance improvements are not achieved by increasing the number of parameters.

Another distinction between NDMs and DDPM is the importantance for NDMs in using the full objective (\ref{eq:elbo}) when training the model. A simplified objective, such as $\Lsimple$ used in DDPM, which measures how well the model predicts injected noise and does not take into account the transformation $F_\vphi$, can cause the collapse of this transformation to $0$. The reason for this is that it becomes trivial to identify the injected noise through $\z_t$.

Finally, unlike conventional diffusion, the generative process of NDMs with learnable transformations depends on the parameters of the forward process. Therefore, in the case of learnable parameters, NDMs do not support conditional generation techniques with classifier guidance \citep{dhariwal2021diffusion}. However, we can utilize alternative approaches \citep{wu2023practical} to enable conditional generation from NDMs, but we will defer this to future research.

\section{Conclusion}
\label{sec:conclusion}

We introduced Neural Diffusion Models (NDMs), a new class of diffusion models that enables defining and learning the general forward noising process. First, we showed how to optimize NDMs using a variational bound in a simulation-free setting. Then, we derived a time-continuous formulation of NDMs allowing for fast and reliable inference and likelihood evaluation using off-the-shelf numerial ODE and SDE solvers. Next, we demonstrated how some existing diffusion models appear as a special cases of NDMs. For NDMs with learnable transformations we studied their utility on standard image generation benchmarks. NDMs significantly outperforms conventional diffusion models in terms of likelihood, achieving state-of-the-art results for ImageNet and CelebA-HQ, and produces samples of comparable or better quality.

\section*{Impact Statement}

This paper presents work whose goal is to advance the field of Machine Learning. There are many potential societal consequences of our work, none which we feel must be specifically highlighted here.

\bibliography{bibliography}

\begin{thebibliography}{57}
\providecommand{\natexlab}[1]{#1}
\providecommand{\url}[1]{\texttt{#1}}
\expandafter\ifx\csname urlstyle\endcsname\relax
  \providecommand{\doi}[1]{doi: #1}\else
  \providecommand{\doi}{doi: \begingroup \urlstyle{rm}\Url}\fi

\bibitem[Albergo \& Vanden-Eijnden(2023)Albergo and Vanden-Eijnden]{albergo2023building}
Albergo, M.~S. and Vanden-Eijnden, E.
\newblock Building normalizing flows with stochastic interpolants.
\newblock In \emph{The Eleventh International Conference on Learning Representations}, 2023.
\newblock URL \url{https://openreview.net/forum?id=li7qeBbCR1t}.

\bibitem[Chen et~al.(2018)Chen, Rubanova, Bettencourt, and Duvenaud]{chen2018neural}
Chen, R.~T., Rubanova, Y., Bettencourt, J., and Duvenaud, D.~K.
\newblock Neural ordinary differential equations.
\newblock \emph{Advances in neural information processing systems}, 31, 2018.

\bibitem[Chen et~al.(2021)Chen, Liu, and Theodorou]{chen2021likelihood}
Chen, T., Liu, G.-H., and Theodorou, E.~A.
\newblock Likelihood training of schr$\backslash$" odinger bridge using forward-backward sdes theory.
\newblock \emph{arXiv preprint arXiv:2110.11291}, 2021.

\bibitem[Creswell et~al.(2018)Creswell, White, Dumoulin, Arulkumaran, Sengupta, and Bharath]{creswell2018generative}
Creswell, A., White, T., Dumoulin, V., Arulkumaran, K., Sengupta, B., and Bharath, A.~A.
\newblock Generative adversarial networks: An overview.
\newblock \emph{IEEE Signal Processing Magazine}, 35\penalty0 (1):\penalty0 53--65, 2018.

\bibitem[Dai et~al.(2017)Dai, Yang, Yang, Cohen, and Salakhutdinov]{dai2017good}
Dai, Z., Yang, Z., Yang, F., Cohen, W.~W., and Salakhutdinov, R.~R.
\newblock Good semi-supervised learning that requires a bad gan.
\newblock \emph{Advances in neural information processing systems}, 30, 2017.

\bibitem[Daras et~al.(2022)Daras, Delbracio, Talebi, Dimakis, and Milanfar]{daras2022soft}
Daras, G., Delbracio, M., Talebi, H., Dimakis, A.~G., and Milanfar, P.
\newblock Soft diffusion: Score matching for general corruptions.
\newblock \emph{arXiv preprint arXiv:2209.05442}, 2022.

\bibitem[De~Bortoli et~al.(2021)De~Bortoli, Thornton, Heng, and Doucet]{de2021diffusion}
De~Bortoli, V., Thornton, J., Heng, J., and Doucet, A.
\newblock Diffusion schr{\"o}dinger bridge with applications to score-based generative modeling.
\newblock \emph{Advances in Neural Information Processing Systems}, 34:\penalty0 17695--17709, 2021.

\bibitem[Deng et~al.(2009)Deng, Dong, Socher, Li, Li, and Fei-Fei]{deng2009imagenet}
Deng, J., Dong, W., Socher, R., Li, L.-J., Li, K., and Fei-Fei, L.
\newblock Imagenet: A large-scale hierarchical image database.
\newblock In \emph{2009 IEEE conference on computer vision and pattern recognition}, pp.\  248--255. Ieee, 2009.

\bibitem[Deng(2012)]{deng2012mnist}
Deng, L.
\newblock The mnist database of handwritten digit images for machine learning research [best of the web].
\newblock \emph{IEEE signal processing magazine}, 29\penalty0 (6):\penalty0 141--142, 2012.

\bibitem[Dhariwal \& Nichol(2021)Dhariwal and Nichol]{dhariwal2021diffusion}
Dhariwal, P. and Nichol, A.
\newblock Diffusion models beat gans on image synthesis.
\newblock \emph{Advances in neural information processing systems}, 34:\penalty0 8780--8794, 2021.

\bibitem[Dormand \& Prince(1980)Dormand and Prince]{dormand1980family}
Dormand, J.~R. and Prince, P.~J.
\newblock A family of embedded runge-kutta formulae.
\newblock \emph{Journal of computational and applied mathematics}, 6\penalty0 (1):\penalty0 19--26, 1980.

\bibitem[Goodfellow et~al.(2014)Goodfellow, Pouget-Abadie, Mirza, Xu, Warde-Farley, Ozair, Courville, and Bengio]{goodfellow2014generative}
Goodfellow, I., Pouget-Abadie, J., Mirza, M., Xu, B., Warde-Farley, D., Ozair, S., Courville, A., and Bengio, Y.
\newblock Generative adversarial nets.
\newblock \emph{Advances in neural information processing systems}, 27, 2014.

\bibitem[Grathwohl et~al.(2019)Grathwohl, Chen, Bettencourt, and Duvenaud]{grathwohl2018scalable}
Grathwohl, W., Chen, R. T.~Q., Bettencourt, J., and Duvenaud, D.
\newblock Scalable reversible generative models with free-form continuous dynamics.
\newblock In \emph{International Conference on Learning Representations}, 2019.
\newblock URL \url{https://openreview.net/forum?id=rJxgknCcK7}.

\bibitem[Gu et~al.(2023)Gu, Zhai, Zhang, Bautista, and Susskind]{gu2023fdm}
Gu, J., Zhai, S., Zhang, Y., Bautista, M.~{\'A}., and Susskind, J.~M.
\newblock f-{DM}: A multi-stage diffusion model via progressive signal transformation.
\newblock In \emph{The Eleventh International Conference on Learning Representations}, 2023.
\newblock URL \url{https://openreview.net/forum?id=iBdwKIsg4m}.

\bibitem[Heusel et~al.(2017)Heusel, Ramsauer, Unterthiner, Nessler, and Hochreiter]{heusel2017gans}
Heusel, M., Ramsauer, H., Unterthiner, T., Nessler, B., and Hochreiter, S.
\newblock Gans trained by a two time-scale update rule converge to a local nash equilibrium.
\newblock \emph{Advances in neural information processing systems}, 30, 2017.

\bibitem[Ho et~al.(2020)Ho, Jain, and Abbeel]{ho2020denoising}
Ho, J., Jain, A., and Abbeel, P.
\newblock Denoising diffusion probabilistic models.
\newblock \emph{Advances in neural information processing systems}, 33:\penalty0 6840--6851, 2020.

\bibitem[Hoogeboom \& Salimans(2023)Hoogeboom and Salimans]{hoogeboom2023blurring}
Hoogeboom, E. and Salimans, T.
\newblock Blurring diffusion models.
\newblock In \emph{The Eleventh International Conference on Learning Representations}, 2023.
\newblock URL \url{https://openreview.net/forum?id=OjDkC57x5sz}.

\bibitem[Karras et~al.(2017)Karras, Aila, Laine, and Lehtinen]{karras2017progressive}
Karras, T., Aila, T., Laine, S., and Lehtinen, J.
\newblock Progressive growing of gans for improved quality, stability, and variation.
\newblock \emph{arXiv preprint arXiv:1710.10196}, 2017.

\bibitem[Kim et~al.(2022)Kim, Na, Kwon, Lee, Kang, and Moon]{kim2022maximum}
Kim, D., Na, B., Kwon, S.~J., Lee, D., Kang, W., and Moon, I.-c.
\newblock Maximum likelihood training of implicit nonlinear diffusion model.
\newblock \emph{Advances in Neural Information Processing Systems}, 35:\penalty0 32270--32284, 2022.

\bibitem[Kingma et~al.(2021)Kingma, Salimans, Poole, and Ho]{kingma2021variational}
Kingma, D., Salimans, T., Poole, B., and Ho, J.
\newblock Variational diffusion models.
\newblock \emph{Advances in neural information processing systems}, 34:\penalty0 21696--21707, 2021.

\bibitem[Kingma \& Welling(2014)Kingma and Welling]{kingma2014auto}
Kingma, D.~P. and Welling, M.
\newblock Auto-encoding variational {B}ayes.
\newblock \emph{International Conference on Learning Representations}, 2014.

\bibitem[Krizhevsky et~al.(2009)Krizhevsky, Hinton, et~al.]{krizhevsky2009learning}
Krizhevsky, A., Hinton, G., et~al.
\newblock Learning multiple layers of features from tiny images.
\newblock 2009.

\bibitem[Lee et~al.(2023)Lee, Kim, and Ye]{lee2023minimizing}
Lee, S., Kim, B., and Ye, J.~C.
\newblock Minimizing trajectory curvature of ode-based generative models.
\newblock In \emph{International Conference on Machine Learning}, pp.\  18957--18973. PMLR, 2023.

\bibitem[Lipman et~al.(2023)Lipman, Chen, Ben-Hamu, Nickel, and Le]{lipman2023flow}
Lipman, Y., Chen, R. T.~Q., Ben-Hamu, H., Nickel, M., and Le, M.
\newblock Flow matching for generative modeling.
\newblock In \emph{The Eleventh International Conference on Learning Representations}, 2023.
\newblock URL \url{https://openreview.net/forum?id=PqvMRDCJT9t}.

\bibitem[Liu et~al.(2022)Liu, Ren, Lin, and Zhao]{liu2022pseudo}
Liu, L., Ren, Y., Lin, Z., and Zhao, Z.
\newblock Pseudo numerical methods for diffusion models on manifolds.
\newblock In \emph{International Conference on Learning Representations}, 2022.
\newblock URL \url{https://openreview.net/forum?id=PlKWVd2yBkY}.

\bibitem[Liu(2022)]{liu2022rectified}
Liu, Q.
\newblock Rectified flow: A marginal preserving approach to optimal transport.
\newblock \emph{arXiv preprint arXiv:2209.14577}, 2022.

\bibitem[Liu et~al.(2023)Liu, Gong, and qiang liu]{liu2023flow}
Liu, X., Gong, C., and qiang liu.
\newblock Flow straight and fast: Learning to generate and transfer data with rectified flow.
\newblock In \emph{The Eleventh International Conference on Learning Representations}, 2023.
\newblock URL \url{https://openreview.net/forum?id=XVjTT1nw5z}.

\bibitem[MacKay(2003)]{mackay2003information}
MacKay, D.~J.
\newblock \emph{Information theory, inference and learning algorithms}.
\newblock Cambridge university press, 2003.

\bibitem[Neklyudov et~al.(2022)Neklyudov, Severo, and Makhzani]{neklyudov2022action}
Neklyudov, K., Severo, D., and Makhzani, A.
\newblock Action matching: A variational method for learning stochastic dynamics from samples.
\newblock \emph{arXiv preprint arXiv:2210.06662}, 2022.

\bibitem[Nichol \& Dhariwal(2021)Nichol and Dhariwal]{nichol2021improved}
Nichol, A.~Q. and Dhariwal, P.
\newblock Improved denoising diffusion probabilistic models.
\newblock In \emph{International conference on machine learning}, pp.\  8162--8171. PMLR, 2021.

\bibitem[Nielsen et~al.(2024)Nielsen, Christensen, Dittadi, and Winther]{nielsen2024diffenc}
Nielsen, B. M.~G., Christensen, A., Dittadi, A., and Winther, O.
\newblock Diffenc: Variational diffusion with a learned encoder.
\newblock In \emph{The Twelfth International Conference on Learning Representations}, 2024.
\newblock URL \url{https://openreview.net/forum?id=8nxy1bQWTG}.

\bibitem[{\O}ksendal \& {\O}ksendal(2003){\O}ksendal and {\O}ksendal]{oksendal2003stochastic}
{\O}ksendal, B. and {\O}ksendal, B.
\newblock \emph{Stochastic differential equations}.
\newblock Springer, 2003.

\bibitem[Papamakarios et~al.(2021)Papamakarios, Nalisnick, Rezende, Mohamed, and Lakshminarayanan]{papamakarios2021normalizing}
Papamakarios, G., Nalisnick, E., Rezende, D.~J., Mohamed, S., and Lakshminarayanan, B.
\newblock Normalizing flows for probabilistic modeling and inference.
\newblock \emph{The Journal of Machine Learning Research}, 22\penalty0 (1):\penalty0 2617--2680, 2021.

\bibitem[Peluchetti()]{peluchettinon}
Peluchetti, S.
\newblock Non-denoising forward-time diffusions.

\bibitem[Popov et~al.(2021)Popov, Vovk, Gogoryan, Sadekova, and Kudinov]{popov2021grad}
Popov, V., Vovk, I., Gogoryan, V., Sadekova, T., and Kudinov, M.
\newblock Grad-tts: A diffusion probabilistic model for text-to-speech.
\newblock In \emph{International Conference on Machine Learning}, pp.\  8599--8608. PMLR, 2021.

\bibitem[Rezende et~al.(2014)Rezende, Mohamed, and Wierstra]{rezende2014stochastic}
Rezende, D.~J., Mohamed, S., and Wierstra, D.
\newblock Stochastic backpropagation and approximate inference in deep generative models.
\newblock In \emph{International conference on machine learning}, pp.\  1278--1286. PMLR, 2014.

\bibitem[Rissanen et~al.(2023)Rissanen, Heinonen, and Solin]{rissanen2023generative}
Rissanen, S., Heinonen, M., and Solin, A.
\newblock Generative modelling with inverse heat dissipation.
\newblock In \emph{The Eleventh International Conference on Learning Representations}, 2023.
\newblock URL \url{https://openreview.net/forum?id=4PJUBT9f2Ol}.

\bibitem[Rombach et~al.(2022)Rombach, Blattmann, Lorenz, Esser, and Ommer]{rombach2022high}
Rombach, R., Blattmann, A., Lorenz, D., Esser, P., and Ommer, B.
\newblock High-resolution image synthesis with latent diffusion models.
\newblock In \emph{Proceedings of the IEEE/CVF conference on computer vision and pattern recognition}, pp.\  10684--10695, 2022.

\bibitem[Saharia et~al.(2021)Saharia, Ho, Chan, Salimans, Fleet, and Norouzi]{saharia2021image}
Saharia, C., Ho, J., Chan, W., Salimans, T., Fleet, D.~J., and Norouzi, M.
\newblock Image super-resolution via iterative refinement.
\newblock \emph{arXiv preprint arXiv:2104.07636}, 2021.

\bibitem[Salimans \& Ho(2022)Salimans and Ho]{salimans2022progressive}
Salimans, T. and Ho, J.
\newblock Progressive distillation for fast sampling of diffusion models.
\newblock In \emph{International Conference on Learning Representations}, 2022.
\newblock URL \url{https://openreview.net/forum?id=TIdIXIpzhoI}.

\bibitem[Singhal et~al.(2023)Singhal, Goldstein, and Ranganath]{singhal2023where}
Singhal, R., Goldstein, M., and Ranganath, R.
\newblock Where to diffuse, how to diffuse, and how to get back: Automated learning for multivariate diffusions.
\newblock In \emph{The Eleventh International Conference on Learning Representations}, 2023.
\newblock URL \url{https://openreview.net/forum?id=osei3IzUia}.

\bibitem[Smale \& Hirsch(1974)Smale and Hirsch]{smale1974differential}
Smale, S. and Hirsch, M.~W.
\newblock \emph{Differential equations, dynamical systems, and linear algebra}, volume~60.
\newblock Elsevier, 1974.

\bibitem[Sohl-Dickstein et~al.(2015)Sohl-Dickstein, Weiss, Maheswaranathan, and Ganguli]{sohl2015deep}
Sohl-Dickstein, J., Weiss, E., Maheswaranathan, N., and Ganguli, S.
\newblock Deep unsupervised learning using nonequilibrium thermodynamics.
\newblock In \emph{International Conference on Machine Learning}, pp.\  2256--2265. PMLR, 2015.

\bibitem[Song et~al.(2021{\natexlab{a}})Song, Meng, and Ermon]{song2021denoising}
Song, J., Meng, C., and Ermon, S.
\newblock Denoising diffusion implicit models.
\newblock In \emph{International Conference on Learning Representations}, 2021{\natexlab{a}}.
\newblock URL \url{https://openreview.net/forum?id=St1giarCHLP}.

\bibitem[Song et~al.(2017)Song, Kim, Nowozin, Ermon, and Kushman]{song2017pixeldefend}
Song, Y., Kim, T., Nowozin, S., Ermon, S., and Kushman, N.
\newblock Pixeldefend: Leveraging generative models to understand and defend against adversarial examples.
\newblock \emph{arXiv preprint arXiv:1710.10766}, 2017.

\bibitem[Song et~al.(2021{\natexlab{b}})Song, Durkan, Murray, and Ermon]{song2021maximum}
Song, Y., Durkan, C., Murray, I., and Ermon, S.
\newblock Maximum likelihood training of score-based diffusion models.
\newblock \emph{Advances in Neural Information Processing Systems}, 34:\penalty0 1415--1428, 2021{\natexlab{b}}.

\bibitem[Song et~al.(2021{\natexlab{c}})Song, Sohl-Dickstein, Kingma, Kumar, Ermon, and Poole]{song2021scorebased}
Song, Y., Sohl-Dickstein, J., Kingma, D.~P., Kumar, A., Ermon, S., and Poole, B.
\newblock Score-based generative modeling through stochastic differential equations.
\newblock In \emph{International Conference on Learning Representations}, 2021{\natexlab{c}}.
\newblock URL \url{https://openreview.net/forum?id=PxTIG12RRHS}.

\bibitem[Tachibana et~al.(2021)Tachibana, Go, Inahara, Katayama, and Watanabe]{tachibana2021taylor}
Tachibana, H., Go, M., Inahara, M., Katayama, Y., and Watanabe, Y.
\newblock It$\backslash$\^{}$\{$o$\}$-taylor sampling scheme for denoising diffusion probabilistic models using ideal derivatives.
\newblock \emph{arXiv preprint arXiv:2112.13339}, 2021.

\bibitem[Tomczak(2022)]{tomczak2022deep}
Tomczak, J.~M.
\newblock \emph{Deep generative modeling}.
\newblock Springer, 2022.

\bibitem[Trippe et~al.(2023)Trippe, Yim, Tischer, Baker, Broderick, Barzilay, and Jaakkola]{trippe2023diffusion}
Trippe, B.~L., Yim, J., Tischer, D., Baker, D., Broderick, T., Barzilay, R., and Jaakkola, T.~S.
\newblock Diffusion probabilistic modeling of protein backbones in {3D} for the motif-scaffolding problem.
\newblock In \emph{The Eleventh International Conference on Learning Representations}, 2023.

\bibitem[Vahdat et~al.(2021)Vahdat, Kreis, and Kautz]{vahdat2021score}
Vahdat, A., Kreis, K., and Kautz, J.
\newblock Score-based generative modeling in latent space.
\newblock \emph{Advances in Neural Information Processing Systems}, 34:\penalty0 11287--11302, 2021.

\bibitem[Van Den~Oord et~al.(2016)Van Den~Oord, Kalchbrenner, and Kavukcuoglu]{van2016pixel}
Van Den~Oord, A., Kalchbrenner, N., and Kavukcuoglu, K.
\newblock Pixel recurrent neural networks.
\newblock In \emph{International conference on machine learning}, pp.\  1747--1756. PMLR, 2016.

\bibitem[Wang et~al.(2021)Wang, Jiao, Xu, Wang, and Yang]{wang2021deep}
Wang, G., Jiao, Y., Xu, Q., Wang, Y., and Yang, C.
\newblock Deep generative learning via schr{\"o}dinger bridge.
\newblock In \emph{International Conference on Machine Learning}, pp.\  10794--10804. PMLR, 2021.

\bibitem[Watson et~al.(2022)Watson, Juergens, Bennett, Trippe, Yim, Eisenach, Ahern, Borst, Ragotte, Milles, et~al.]{watson2022broadly}
Watson, J.~L., Juergens, D., Bennett, N.~R., Trippe, B.~L., Yim, J., Eisenach, H.~E., Ahern, W., Borst, A.~J., Ragotte, R.~J., Milles, L.~F., et~al.
\newblock Broadly applicable and accurate protein design by integrating structure prediction networks and diffusion generative models.
\newblock \emph{bioRxiv}, pp.\  2022--12, 2022.

\bibitem[Wu et~al.(2023)Wu, Trippe, Naesseth, Blei, and Cunningham]{wu2023practical}
Wu, L., Trippe, B.~L., Naesseth, C.~A., Blei, D.~M., and Cunningham, J.~P.
\newblock Practical and asymptotically exact conditional sampling in diffusion models.
\newblock \emph{arXiv preprint arXiv:2306.17775}, 2023.

\bibitem[Xiao et~al.(2021)Xiao, Kreis, and Vahdat]{xiao2021tackling}
Xiao, Z., Kreis, K., and Vahdat, A.
\newblock Tackling the generative learning trilemma with denoising diffusion {GANs}.
\newblock \emph{arXiv preprint arXiv:2112.07804}, 2021.

\bibitem[Yang et~al.(2022)Yang, Zhang, Song, Hong, Xu, Zhao, Shao, Zhang, Cui, and Yang]{yang2022diffusion}
Yang, L., Zhang, Z., Song, Y., Hong, S., Xu, R., Zhao, Y., Shao, Y., Zhang, W., Cui, B., and Yang, M.-H.
\newblock Diffusion models: A comprehensive survey of methods and applications.
\newblock \emph{arXiv preprint arXiv:2209.00796}, 2022.

\end{thebibliography}
\bibliographystyle{icml2024}

\newpage
\appendix
\onecolumn

\section{Derivations and proofs}
\label{app:method}

\subsection{Forward posterior}
\label{app:posterior}

First, we rewrite the marginal distribution (\ref{eq:ndm_marginal}) in terms of standard normally distributed $\veps_t$, $\veps_s$ for $s$ and $t$, where $s < t$:
\begin{align}
    \label{eq:marginal_sample_zt}
    \z_t &= \a_t F_\vphi (\x, t) + \s_t \veps_t, \\
    \z_s &= \a_s F_\vphi (\x, s) + \s_s \veps_s.
\end{align}

Next, we constructively introduce the posterior distribution $q_\vphi(\z_s|\z_t, \x)$. To sample $\z_s$ given $\z_t$ and $\x$ while preserving the correct marginal distribution $q_\vphi(\z_s|\x)$, we can combine the noise $\veps_t$ with additional noise $\tilde{\veps}_{s|t}$ as follows:
\begin{align}
    \z_s = \a_s F_\vphi (\x, s) + \sqrt{\s_s^2 - \ts_{s|t}^2} \veps_t + \ts_{s|t} \tilde{\veps}_{s|t}.
    \label{eq:posterior_sample_zs_eps}
\end{align}

The samples $\z_s$ follow a (conditional) normal distribution. By marginalizing $\veps_t$ and $\tilde{\veps}_{s|t}$, we obtain a normal distribution with mean $\a_s F\vphi (\x, s)$ and variance $\s_s^2 - \ts_{s|t}^2 + \ts_{s|t}^2 = \s_s^2$. Therefore, this sampling procedure satisfies $q_\vphi(\z_s|\x) = \int q_\vphi(\z_t|\x) q_\vphi(\z_s|\z_t, \x) d \z_t$.

The equation (\ref{eq:posterior_sample_zs_eps}) relies on $\veps_t$, which we do not have explicit access to. However, once we know $\z_t$ and $\x$, we can calculate it from (\ref{eq:marginal_sample_zt}) as $\veps_t = \frac{\z_t - \a_t F_\vphi (\x, t)}{\s_t}$ and substitute it in (\ref{eq:posterior_sample_zs_eps}):
\begin{align}
    \z_s = \a_s F_\vphi (\x, s) + \frac{\sqrt{\s_s^2 - \ts_{s|t}^2}}{\s_t} \Big( \z_t - \a_t F_\vphi (\x, t) \Big) + \ts_{s|t} \tilde{\veps}_{s|t}.
\end{align}

Using this constructive definition, we obtain the posterior distribution (\ref{eq:ndm_posterior}).

\subsection{Objective}
\label{app:objective}

To calculate the diffusion term $\Ldiff$ (\ref{eq:ndm_kl}) of the objective, we need to compute the KL divergence between the forward posterior distribution $q_\vphi(\mathbf{z}_s|\mathbf{z}_t, \mathbf{x})$ and the reverse distribution $p_{\theta}(\mathbf{z}_s|\mathbf{z}_t)$. Since we use parameterization $p_{\theta} (\z_s|\z_t) = q_\vphi(\z_s|\z_t, \hx_{\theta}(\z_t, t))$, both of these distributions are normal distributions with the same variance, so we can evaluate the KL divergence between them analytically as follows:
\begin{align}
    \KL \Big(  q_\vphi(\z_s|\z_t, \x) || & p_{\theta} (\z_s|\z_t) \Big) = \nonumber \\ 
    ~
    = \frac{1}{2 \ts_{s|t}^2} \Bigg\| 
    & \a_s F_\vphi (\x, s) + \frac{\sqrt{\s_s^2 - \ts_{s|t}^2}}{\s_t} \Big( \cancel{\z_t} - \a_t F_\vphi (\x, t) \Big) - \nonumber \\
    & \a_s F_\vphi(\hx_{\theta}(\z_t, t), s) - \frac{\sqrt{\s_s^2 - \ts_{s|t}^2}}{\s_t} \Big( \cancel{\z_t} - \a_t F_\vphi(\hx_{\theta}(\z_t, t), t) \Big)
    \Bigg\|_2^2 \\
    ~
    = \frac{1}{2 \ts_{s|t}^2} \Bigg\| 
    & \a_s \Big( F_\vphi(\x, s) - F_\vphi(\hx_{\theta}(\z_t, t), s) \Big) + \nonumber \\
    & \frac{\sqrt{\s_s^2 - \ts_{s|t}^2}}{\s_t} \a_t \Big( F_\vphi(\hx_{\theta}(\z_t, t), t) - F_\vphi(\x, t) \Big)
    \Bigg\|_2^2.
\end{align}

With a learnable transformation $F_\vphi$, the term $\Lprior$ becomes dependent on the parameters $\vphi$, necessitating its optimization during training. We can compute the prior term as follows:
\begin{align}
    \KL \Big(  q_\vphi(\z_T|\x) || p (\z_T) \Big) 
    &= \frac{1}{2} \left[ \log \frac{|I|}{|\s_T^2 I|} - d + Tr\{I^{-1} \s_t^2 I\} + \Big\| 0 - \a_T F_\vphi(\x, T)\Big \|_2^2 \right] \\
    &= \frac{1}{2} \left[ - d \log \s_T^2 - d + d \s_T^2 + \Big\| \a_T F_\vphi(\x, T)\Big \|_2^2 \right] \\
    &= \frac{1}{2} \left[ d \Big( \s_T^2 - \log \s_T^2 - 1 \Big) + \a_T^2 \Big\| F_\vphi(\x, T)\Big \|_2^2 \right].
\end{align}
Here, $d$ represents the dimensionality of the data space.

\subsection{Reverse SDE and ODE}
\label{app:sde}

As discussed in Section \ref{sec:cont_ndm}, when the number of steps, denoted as $T$, tends to infinity for NDM, we can switch to continuous time.

In the discrete time setting, we define the time step as $t \in [0, 1, \dots, T]$. In the continuous time setting, we utilize the unit interval, denoting time as $t \in [0, 1]$. Nevertheless, for the sake of notational simplicity in this and subsequent sections, we will consider the discrete time to also lie within the unit interval, with $t \in [\frac{0}{T}, \frac{1}{T}, \dots, \frac{T}{T}]$.

To derive the stochastic differential equation (SDE) for the reverse process $p_\theta(\z_s|\z_t)$ in NDM, we first obtain an SDE that depends on the data point $\x$ and whose solution corresponds to the posterior distribution $q_\vphi(\z_{t - \D t}|\z_t, \x)$. By defining $p_\theta(\z_s|\z_t)$ through $q_\vphi(\z_s|\z_t, \x)$ with the prediction $\hx_{\theta}(\z_t, t)$ instead of $\x$, we can subsequently replace the prediction $\hx_{\theta}(\z_t, t)$ and derive the SDE for the reverse process.

We constructively derive the SDE for the posteriors $q_\vphi(\z_s|\z_t, \x)$. First, let us consider the following auxiliary SDE with backward time flow:
\begin{align}
    d \veps_t = \frac{\dot{\mnu}_t}{2} \veps_t d t + \sqrt{\dot{\mnu}_t} d \w.
    \label{eq:eps_sde}
\end{align}

It is straightforward to show that the solution to this SDE corresponds to the following distribution:
\begin{align}
    q(\veps_s|\veps_t) = \N(\veps_s; \sqrt{1 - \bar{\s}_{s|t}^2} \veps_t; \bar{\s}_{s|t}^2 I), \quad \text{where} \quad \bar{\s}_{s|t}^2 = 1 - e^{\mnu_s - \mnu_t}.
    \label{eq:eps_distribution}
\end{align}

To derive the SDE for the posteriors $q_\vphi(\z_s|\z_t, \x)$, we can apply the following function to both the SDE (\ref{eq:eps_sde}) and the distribution (\ref{eq:eps_distribution}):
\begin{align}
    G(\x, \veps_t, t) = \a_t F_\vphi (\x, t) + \s_t \veps_t
\end{align}
Note that after applying the function $G$, the distribution (\ref{eq:eps_distribution}) matches the posterior distribution $q_\vphi(\z_s|\z_t, \x)$. Therefore, the desired SDE for $q_\vphi(\z_s|\z_t, \x)$ is obtained by transforming the SDE (\ref{eq:eps_sde}) using Ito's formula \citep{oksendal2003stochastic}:
\begin{align}
    d \z_t
    &= \left[ \frac{\d G(\x, \veps, t)}{\d t} \Big\vert_{\veps=\veps_t} + \frac{\dot{\mnu}_t}{2} \frac{\d G(\x, \veps_t, t)}{\d \veps_t} \veps_t - \frac{\dot{\mnu}_t}{2} \frac{\d^2 G(\x, \veps_t, t)}{\d \veps_t^2} \right] d t + \sqrt{\dot{\mnu}_t} \frac{\d G(\x, \veps_t, t)}{\d \veps_t} d \w \\
    &= \left[ \dot{\a}_t F_\vphi (\x, t) + \a_t \dot{F}_\vphi (\x, t) + \dot{\s}_t \veps_t + \frac{\dot{\mnu}_t}{2} \s_t \veps_t \right] d t + \sqrt{\dot{\mnu}_t} \s_t d \w \\
    &= \left[ \frac{\dot{\a}_t}{\a_t} (\z_t - \s_t \veps_t) + \a_t \dot{F}_\vphi (\x, t) + \dot{\s}_t \veps_t + \frac{\dot{\mnu}_t}{2} \s_t \veps_t \right] d t + \sqrt{\dot{\mnu}_t} \s_t d \w \\
    &= \left[ \a_t \dot{F}_\vphi (\x, t) + \frac{\d \log \a_t}{\d t} \z_t - \frac{1}{2} \left( \frac{\d \s_t^2}{\d t} - 2 \frac{\d \log \a_t}{\d t} \s_t^2 + \dot{\mnu}_t \s_t^2 \right) \left( - \frac{\veps_t}{\s_t} \right) \right] d t + \sqrt{\dot{\mnu}_t} \s_t d \w \\
    &= \left[ \a_t \dot{F}_\vphi (\x, t) + r(t) \z_t - \frac{1}{2} \left( \frac{\d \s_t^2}{\d t} - 2 r(t) \s_t^2 + \dot{\mnu}_t \s_t^2 \right) s(\x, \z_t, t) \right] d t + \sqrt{\dot{\mnu}_t} \s_t d \w, \label{eq:forward_sde} \\
    & \textrm{where} \quad r(t) = \frac{\d \log \a_t}{\d t} \quad \textrm{and} \quad s(\x, \z_t, t) = \frac{\a_t F_\vphi(\x, t) - \z_t}{\s_t^2}
\end{align}

To obtain the SDE for the reverse process, we can substitute the prediction $\hx_{\theta}(\z_t, t)$ instead of $\x$. This substitution yields the SDE (\ref{eq:sde}):
\begin{align}
    d \z_t 
    &= \left[ \a_t \dot{F}_\vphi (\hx_{\theta}(\z_t, t), t) + r(t) \z_t - \frac{1}{2} \left( \frac{\d \s_t^2}{\d t} - 2 r(t) \s_t^2 + \dot{\mnu}_t \s_t^2 \right) s_\theta(\z_t, t) \right] d t + \sqrt{\dot{\mnu}_t} \s_t d \w, \label{eq:reverse_sde} \\
    & \textrm{where} \quad s_\theta(\z_t, t) = \frac{\a_t F_\vphi(\hx_{\theta}(\z_t, t), t) - \z_t}{\s_t^2}
\end{align}
As discussed earlier, we can leverage the Jacobian-Vector Product (JVP) trick \citep{smale1974differential} to calculate $\dot{F}_\vphi$.

In the case where $\mnu_t$ is a constant, the dynamics become deterministic and can be described by ordinary differential equations (ODEs). In our experiments, we utilize these ODEs to model the generative process as a continuous normalizing flow \citep{chen2018neural, grathwohl2018scalable} and estimate densities.

\subsection{Continuous time objective}
\label{app:cont_objective}

When we switch to continuous time, the discrete objective (\ref{eq:ndm_kl}) transforms from finite sum of KL divergances into integral, which we can easily derive as soon as we have access to both stochastic differential equations associated with the forward process (\ref{eq:forward_sde}) and with the rewerse process (\ref{eq:reverse_sde}). In continuous time the diffusion term $\Ldiff$ (\ref{eq:ldiff_cont}) is equal to:

\begin{align}
    \Ldiff = 
    ~
    \int_0^1
    \frac{1}{g^2(t)} \Bigg\| &
    \a_t \Big( \dot{F}_\vphi (\x, t) - \dot{F}_\vphi (\hx_{\theta}(\z_t, t), t) \Big) + \nonumber \\
    & \frac{1}{2} \left( \frac{\d \s_t^2}{\d t} - 2 r(t) \s_t^2 + g^2(t) \right) 
    \Big( s(\x, \z_t, t) - s(\hx_{\theta}(\z_t, t), \z_t, t) \Big)
    \Bigg\|_2^2 d t, \nonumber \\
    ~
    \textrm{where} \quad r(t) = \frac{\d \log \a_t}{\d t}, & \quad g^2(t) = \dot{\mnu}_t \s_t^2 \quad \textrm{and} \quad s(\x, \z_t, t) = \frac{\a_t F_\vphi(\x, t) - \z_t}{\s_t^2}.
\end{align}

As we can see, these equation contains $\dot{F}_\vphi$ as a component. In general, we do not have explicit access to the time derivative of the forward transformation $F_\vphi$. However, we will focus on cases where the forward transformation is differentiable. By utilizing automatic differentiation tools, we can calculate the time derivatives of $F_\vphi$. Nevertheless, when $\x$ is fixed, the function $F_\vphi(\x, \cdot)$ becomes a scalar-to-vector function. To compute its time derivative using simple backpropagation, we would need to execute it for all outputs of $F_\vphi$, resulting in quadratic computational complexity. Fortunately, there exists a more efficient method to obtain the time derivative, the Jacobian-Vector Product trick \citep{smale1974differential}. The Jacobian of the transformation function with $\x$ fixed is represented as a column matrix. Therefore, by computing the product of the Jacobian with a one-dimensional vector, we can obtain a vector of time derivatives.

\section{Connections with other works}
\label{app:related}

\begin{table}[!t]
\caption{Summary of existing diffusion models as instances of Neural Diffusion Models (NDM).}
\label{tab:generalization_ext}
\centering
\scriptsize
\begin{tabular}{lccl}
\toprule


Model 
& 
Distribution $q(\z_t|x)$ 
& 
NDM's $F(\x, t)$
& 
Comment
\\
\midrule


\makecell[l]{
    DDPM \citep{ho2020denoising} / \\
    DDIM \citep{song2021denoising}
}
& 
$\N\Big(\z_t; \a_t \x, \s^2_t I \Big)$ 
& 
$\x$
& 

\\
\\


\makecell[l]{
    Flow Matching OT \\
    \citep{lipman2023flow}
}
& 
$\N\Big(\z_t; \a_t \x, \s^2_t I \Big)$ 
& 
$\x$
& 
\makecell[l]{
    $\a_t = t$, \\
    $\s_t = 1 - (1 - \s_{\mathrm{min}})t$ 
}
\\
\\


VDM \citep{kingma2021variational}
& 
$\N\Big(\z_t; \a_t \x, \s^2_t I \Big)$ 
& 
$\x$
& 
\makecell[l]{
    $\a^2_t = \mathrm{sigmoid}(-\gamma_{\eta}(t))$, \\
    $\s^2_t = \mathrm{sigmoid}(\gamma_{\eta}(t))$ 
}
\\
\\


IHDM \citep{rissanen2023generative}
&
$\N\Big(\z_t; V e^{-\Lambda t} V^T \x, \s^2 I \Big)$ 
&
$V e^{-\Lambda t} V^T \x$
& 
\makecell[l]{
    $\a_t = 1$, $\s_t = \s$, \\
    $\s$ is fixed
}
\\
\\


\makecell[l]{
    Blurring Diffusion \\
    \citep{hoogeboom2023blurring}
}
&
$\N\Big(\z_t; \a_t e^{-\Lambda t} V^T \x, \s_t^2 I \Big)$ 
&
$e^{-\Lambda t} V^T \x$
& 
$p(x|z_0) = \N\Big( x; a V z_0, \s \Big)$
\\
\\


\makecell[l]{
    Soft Diffusion \\
    \citep{daras2022soft}
}
&
$\N\Big(\z_t; C_t \x, s_t^2 I \Big)$ 
& 
$C_t \x$
&
$\a_t = 1$, $\s^2_t = s^2_t$
\\
\\


LSGM \citep{vahdat2021score}
&
$\N\Big(\z_t; \a_t E(\x), \s_t^2 I \Big)$ 
&
$E(\x)$
& 
$p(x|z_0) = \N\Big( x; a D(z_0), \s^2 \Big)$
\\
\\


f-DM \citep{gu2023fdm}
&
$\N\Big(\z_t; \a_t \x_t, \s_t^2 I \Big)$
&
\makecell[c]{
    $\x_t = \frac{(t - \tau_k) \hat{x}^k + (\tau_{k+1} - t) x^k}{\tau_{k+1} - \tau_k}$, \\
    where $\tau_k \leq t < \tau_{k+1}$
}
& 
\makecell[l]{
    $
    \begin{aligned}
        x^k &= f_{0:k}(x) \\
        \hat{x}^k &= \begin{cases}
            g_k(f_{k+1}(x^k)), & \text{if $k<K$},\\
            x^k, & \text{if $k=K$}.
        \end{cases}
    \end{aligned}
    $
}
\\


\bottomrule
\end{tabular}
\end{table}

We introduce NDMs as a comprehensive framework that generalises various existing approaches. Here we provide Table \ref{tab:generalization_ext} which is an extended version of Table \ref{tab:generalization}, that demonstrates how existing approaches appear as a spatial cases of NDMs.

We also provide an extended discussion on the connection between NDM and other related works.

\subsection{Diffusion in latent space}

The concept of a learnable forward process is not entirely new. In some sense models that run a diffusion process in the latent space of a VAE \citep{vahdat2021score, rombach2022high}, a hierarchical VAE \citep{gu2023fdm}, or a Flow model \citep{kim2022maximum} can be viewed as diffusion models with a learnable forward process. These models optimize the mapping to the latent space. Consequently, projecting the diffusion generative dynamic from the latent to the data space introduces a novel, nonlinear, and learnable generative dynamic. However, these models still rely on conventional diffusion in the latent space.

Additionally, these models can be viewed as spatial cases of NDM with a specific choice of the transformation $F_\vphi(\x, t)$. For example, $F_\vphi$ might be selected as the VAE's time independent encoder in the case of \citep{vahdat2021score} or the time independent Flow model in the case of \citep{kim2022maximum}.

\subsection{Schrödinger Bridges}

Another line of works \citep{de2021diffusion, wang2021deep, peluchettinon, chen2021likelihood} are approaches based on Schrödinger Bridge theory. While such approaches allow learning forward transformations, in contrast to NDM, these approaches are not simulation-free. In Schrödinger Bridge models, we typically lack direct access to the distribution $q(\z_t|\x)$. Consequently, to sample the latent variable $\mathbf{z}_t$ in training time, we must simulate the full stochastic process, such as the stochastic differential equations. This characteristic makes Schrödinger Bridge models expensive in training and not simulation-free. 

In contrast, NDM framework, by design, has access to $q(\z_t|\x)$. Thus, with NDM, when training a model with $T$ time steps, there is no need to propagate $F_\vphi$ for $T$ times at each step of the training procedure. Instead, the NDM framework enables sampling of the intermediate latent variables $\z_t$ directly from the distribution $q(\z_t|\x)$. Therefore, we can maintain the training paradigm outlined in Section \ref{sec:background}. Instead of computing all $T$ KL divergences for each time step, we can approximate the objective using the Monte Carlo method by calculating just one KL divergence for a uniformly sampled time step $t \in [1; T]$, as described in Algorithm \ref{alg:training}.
    
This approach allows us to train the model with batches of shape $[batch\_size, d]$ rather than $[batch\_size, T, d]$. Consequently, NDM can leverage larger batch sizes and use just one call of $F_\vphi$ for inferring latent variables $\z_t$.

\subsection{Stochastic Interpolants}

\citet{albergo2023building} proposed a Stochastic Interpolant approach, which provide more flexibility then conventional diffusion models in defining and even learning of the forward process. While we find stochastic interpolants intriguing and promising, as well as related to our work, these methods differ significantly.

Firstly, stochastic interpolants represent an approach to learning continuous-time deterministic generative dynamics, whereas NDM learns stochastic dynamics in either discrete or continuous time, which can subsequently may be transformed into a deterministic process.

Secondly, in NDM, the model is trained by optimizing the variational bound on the likelihood, while Stochastic Interpolants are trained by optimizing the generalization of the Flow Matching objective \citep{lipman2023flow}.

Lastly, NDM joint learns both the forward and reverse processes by optimizing the likelihood, whereas stochastic interpolants learn the generative process with a fixed forward process. \citet{albergo2023building} demonstrate the possibility of constructing an optimization procedure for the forward process through a max-min game to solve a dynamic optimal transport problem. However, the purpose of this optimization differs from that of NDM.

Moreover, max-min optimization, as employed in Stochastic Interpolants, is notably less stable compared to min-min optimization in NDM. Additionally, Stochastic Interpolants do not present experimental results for the optimization of the forward process.

\subsection{DiffEnc}

In concurrent work, \citet{nielsen2024diffenc} introduced DiffEnc. DiffEnc also proposes to add a time-dependent transformation to the data in the diffusion model. However, there are some distinctions between these two methods. Firstly, in NDM, we parameterize the reverse process by predicting the data point $\x$, while in DiffEnc, they predict the transformed data point $F_\vphi(\x, t)$.

Secondly, in NDM, we employ a Signal-to-Noise Ratio (SNR) schedule for noise injection from DDPM \citep{ho2020denoising} and a straightforward parameterization of the model $\hx_\theta(\z_t, t)$ through predicting the injected epsilon, as detailed in Appendix \ref{app:parameterization}. Simultaneously, in DiffEnc, the authors use a learnable SNR schedule \citep{kingma2021variational} and a v-parameterization \citep{salimans2022progressive} of $\hx_\theta(\z_t, t)$.

Finally, DiffEnc utilizes approximations of the time derivatives of data transformations $F_\vphi$, while in the NDM framework, we propose calculating exact time derivatives using Jacobian-Vector Products.

\section{Implementation details}
\label{app:implementation}

All our experiments were conducted using synthetic 2D datasets and image datasets: MNIST \citep{deng2012mnist}, CIFAR-10 \citep{krizhevsky2009learning}, downsampled ImageNet \citep{deng2009imagenet, van2016pixel} and CelebA-HQ-256 \citep{karras2017progressive}. For CIFAR-10 and ImageNet datasets we applied center cropping and resizing. For synthetic data, we employed a 5-layer MLP with 512 neurons in each layer, while for the images, we utilized the U-Net architecture from \citet{dhariwal2021diffusion}. In our experiments both the DDPM and NDM approaches were trained on identical architectures, with the same hyper-parameters and for the same number of epochs. The hyper-parameters are presented in Table \ref{tab:hyper_parameters}. In experiment where we report results for the continuous time models we use importance sampling of time \citep{song2021maximum} instead of uniform sampling.

We trained models using the Adam optimizer, setting the following parameters: $\beta_1 = 0.9$, $\beta_2 = 0.999$, weight decay of $0.0$, and $\veps = 10^{-8}$. To facilitate the training process, we employed a polynomial decay learning rate schedule, which includes a warm-up phase for a specified number of training steps. During the warm-up phase, the learning rate is linearly increased from $10^{-8}$ to the peak learning rate. Once the peak learning rate is reached, the learning rate is linearly decayed to $10^{-8}$ until the final training step. The training was performed using Tesla V100 GPUs.

\begin{table}[!t]
\caption{Training hyper-parameters.}
\label{tab:hyper_parameters}
\centering
\begin{tabular}{lccc}
\toprule
 & \textbf{CIFAR-10} & \textbf{ImageNet 32} & \textbf{ImageNet 64} \\ \midrule
Channels & 256 & 256 & 192 \\
Depth & 2 & 3 & 3 \\
Channels multipliers & 1,2,2,2 & 1,2,2,2 & 1,2,3,4 \\
Heads & 4 & 4 & 4 \\
Heads Channels & 64 & 64 & 64 \\
Attention resolution & 16 & 16,8 & 32,16,8 \\
Dropout & 0.0 & 0.0 & 0.0 \\
Effective Batch size & 256 & 1024 & 2048 \\
GPUs & 2 & 4 & 16 \\
Epochs & 1000 & 200 & 250 \\
Iterations & 391k & 250k & 157k \\
Learning Rate & 4e-4 & 1e-4 & 1e-4 \\
Learning Rate Scheduler & Polynomial & Polynomial & Constant \\
Warmup Steps & 45k & 20k & - \\
\bottomrule
\end{tabular}
\end{table}

\subsection{Dequantization}

When reporting negative log-likelihood, we dequantize using the standard uniform dequantization. We report an importance-weighted estimate using
\begin{align}
    \log \frac{1}{K} \sum_{k=1}^K p_{\theta}(\x + u_k), \quad \text{where} \quad u_k \sim \mathcal{U}(0, 1),
\end{align}
with $\x \in [0, \dots, 255]$.

\subsection{Parameterization}
\label{app:parameterization}

In order to simplify the derivations above, we have utilized the notation $\hx_{\theta}(\z_t, t)$ to represent the prediction of the reverse process. However, prior research has shown that predicting the injected noise $\veps_t$ can lead to improved results \citep{ho2020denoising, nichol2021improved, dhariwal2021diffusion}. Therefore, in all the experiments, we opt for the following parameterization:
\begin{align}
    \hx_{\theta}(\z_t, t) = \frac{\z_t - \s_t \hat{\veps}_{\theta}(\z_t, t)}{\a_t}.
    \label{eq:reparam_eps}
\end{align}
It is worth noting that with this parameterization, $\hat{\veps}_{\theta}(\z_t, t)$ does not necessarily approximate the true injected noise $\veps_t$, since this reparameterization does not account for the transformation $F_\vphi$. We believe that better parameterizations may exist for NDM, but we leave this for future research.

Furthermore, we restrict the transformation $F_\vphi$ to an identity transformation for $t=0$ through the following construction:
\begin{align}
    F_\vphi(\x, t) = (1 - t) \x + t \bar{F}_\vphi(\x, t).
    \label{eq:reparam_transform}
\end{align}
This ensures that $q(\z_0|\x) \approx \delta(\z_0 - \x)$, and thus also removes the need to optimize the reconstruction term $\Lrec$.

Finally, to ensure consistency with \citet{ho2020denoising} we use $\ts_{s|t}^2 = \left( \s_t^2 - \frac{\a_t^2}{\a_s^2} \s_s^2 \right) \frac{\s_s^2}{\s_t^2}$ for the forward process (\ref{eq:ndm_posterior}). This choice of $\ts_{s|t}^2$ guaranties consistency between the NDM and DDPM forward processes. For $\a_t$ and $\s_t^2$ we use the DDPM schedule of noise injection.

\subsection{Diffusion in latent space}

For experiment with diffusion in the latent space of VAE on CelebA-HQ-256, we followed LSGM \citep{vahdat2021score} experiment setup. The only difference between LSGM baseline and our model is that we utilize learnable transformations $F_\vphi$ according to NDMs framework. We apply the same hyperparameters, as LSGM.

\section{Additional results}
\label{app:additional}

\subsection{Additional evaluation}

\begin{table}[!t]
\caption{Performance comparison the DDPM and NDM on CIFAR-10 and ImageNet 32 datasets with different numbers of steps. We report the performance with same hyperparameters and neural networks on both models to quantify the effect of learnable transformation in fair setting. We provide likelihood (bits/dim) and negative ELBO. Additionally for CIFAR-10 and ImageNet 32 we provide FID score. Boldface numbers represent the best performance. NDM consistently outperforms in terms of NLL and NELBO with comparable sample quality to DDPM on all datasets.}
\label{tab:results_ext}
\centering
\begin{tabular}{lllrrrrrrr}
\toprule
                                         &                             &                          & \multicolumn{3}{c}{\textbf{CIFAR-10}}                                                                                & \multicolumn{1}{c}{}     & \multicolumn{3}{c}{\textbf{ImageNet 32}}                                                                             \\ \cmidrule(r){4-6} \cmidrule(r){8-10}
Steps                                    & Model                       &                          & \multicolumn{1}{c}{NLL $\downarrow$} & \multicolumn{1}{c}{NELBO $\downarrow$} & \multicolumn{1}{c}{FID $\downarrow$} & \multicolumn{1}{c}{}     & \multicolumn{1}{c}{NLL $\downarrow$} & \multicolumn{1}{c}{NELBO $\downarrow$} & \multicolumn{1}{c}{FID $\downarrow$} \\ \cmidrule(r){1-2} \cmidrule(r){4-6} \cmidrule(r){8-10}
                                         & DDPM                        &                          & 3.11                                 & 3.18                                   & \textbf{11.44}                                 &                          & 3.89                                 & 3.95                                   & \textbf{16.18}                                 \\
\multirow{-2}{*}{1000}                   & \cellcolor[HTML]{EFEFEF}NDM & \cellcolor[HTML]{EFEFEF} & \cellcolor[HTML]{EFEFEF}\textbf{3.02}         & \cellcolor[HTML]{EFEFEF}\textbf{3.03}           & \cellcolor[HTML]{EFEFEF}11.82         & \cellcolor[HTML]{EFEFEF} & \cellcolor[HTML]{EFEFEF}\textbf{3.79}         & \cellcolor[HTML]{EFEFEF}\textbf{3.82}           & \cellcolor[HTML]{EFEFEF}17.02         \\ \cmidrule(r){1-2} \cmidrule(r){4-6} \cmidrule(r){8-10}
                                         & DDPM                        &                          & 3.31                                 & 3.38                                   & \textbf{11.78}                                 &                          & 4.14                                 & 4.23                                   & \textbf{16.66}                                 \\
\multirow{-2}{*}{100}                    & \cellcolor[HTML]{EFEFEF}NDM & \cellcolor[HTML]{EFEFEF} & \cellcolor[HTML]{EFEFEF}\textbf{3.05}         & \cellcolor[HTML]{EFEFEF}\textbf{3.12}           & \cellcolor[HTML]{EFEFEF}11.98         & \cellcolor[HTML]{EFEFEF} & \cellcolor[HTML]{EFEFEF}\textbf{3.83}         & \cellcolor[HTML]{EFEFEF}\textbf{3.92}           & \cellcolor[HTML]{EFEFEF}17.74         \\ \cmidrule(r){1-2} \cmidrule(r){4-6} \cmidrule(r){8-10}
                                         & DDPM                        &                          & 3.49                                 & 3.57                                   & 13.22                                 &                          & 4.37                                 & 4.47                                   & \textbf{18.70}                                 \\
\multirow{-2}{*}{50}                     & \cellcolor[HTML]{EFEFEF}NDM & \cellcolor[HTML]{EFEFEF} & \cellcolor[HTML]{EFEFEF}\textbf{3.22}         & \cellcolor[HTML]{EFEFEF}\textbf{3.30}           & \cellcolor[HTML]{EFEFEF}\textbf{13.15}         & \cellcolor[HTML]{EFEFEF} & \cellcolor[HTML]{EFEFEF}\textbf{4.05}         & \cellcolor[HTML]{EFEFEF}\textbf{4.14}           & \cellcolor[HTML]{EFEFEF}18.93         \\ \cmidrule(r){1-2} \cmidrule(r){4-6} \cmidrule(r){8-10}
                                         & DDPM                        &                          & 5.02                                 & 5.13                                   & 37.83                                 &                          & 6.28                                 & 6.42                                   & 53.51                                 \\
\multirow{-2}{*}{10}                     & \cellcolor[HTML]{EFEFEF}NDM & \cellcolor[HTML]{EFEFEF} & \cellcolor[HTML]{EFEFEF}\textbf{4.63}         & \cellcolor[HTML]{EFEFEF}\textbf{4.74}           & \cellcolor[HTML]{EFEFEF}\textbf{31.56}         & \cellcolor[HTML]{EFEFEF} & \cellcolor[HTML]{EFEFEF}\textbf{5.81}         & \cellcolor[HTML]{EFEFEF}\textbf{5.94}           & \cellcolor[HTML]{EFEFEF}\textbf{45.38}         \\ \midrule
                                         & DDPM                        &                          & 3.38                                 & 3.45                                   & \textbf{12.29}                                 &                          & 4.23                                 & 4.32                                   & \textbf{17.49}                                 \\
\multirow{-2}{*}{1000 $\rightarrow$ 100} & \cellcolor[HTML]{EFEFEF}NDM & \cellcolor[HTML]{EFEFEF} & \cellcolor[HTML]{EFEFEF}\textbf{3.30}         & \cellcolor[HTML]{EFEFEF}\textbf{3.37}           & \cellcolor[HTML]{EFEFEF}12.70         & \cellcolor[HTML]{EFEFEF} & \cellcolor[HTML]{EFEFEF}\textbf{4.15}         & \cellcolor[HTML]{EFEFEF}\textbf{4.23}           & \cellcolor[HTML]{EFEFEF}18.48         \\ \cmidrule(r){1-2} \cmidrule(r){4-6} \cmidrule(r){8-10}
                                         & DDPM                        &                          & 4.08                                 & 4.17                                   & \textbf{15.24}                                 &                          & 5.10                                 & 5.21                                   & \textbf{20.09}                                 \\
\multirow{-2}{*}{1000 $\rightarrow$ 50}  & \cellcolor[HTML]{EFEFEF}NDM & \cellcolor[HTML]{EFEFEF} & \cellcolor[HTML]{EFEFEF}\textbf{3.98}         & \cellcolor[HTML]{EFEFEF}\textbf{4.07}           & \cellcolor[HTML]{EFEFEF}16.83         & \cellcolor[HTML]{EFEFEF} & \cellcolor[HTML]{EFEFEF}\textbf{5.00}         & \cellcolor[HTML]{EFEFEF}\textbf{5.10}           & \cellcolor[HTML]{EFEFEF}21.11         \\ \cmidrule(r){1-2} \cmidrule(r){4-6} \cmidrule(r){8-10}
                                         & DDPM                        &                          & 8.78                                 & 8.98                                   & \textbf{43.85}                                 &                          & 10.99                                 & 11.23                                   & \textbf{58.35}                                 \\
\multirow{-2}{*}{1000 $\rightarrow$ 10}  & \cellcolor[HTML]{EFEFEF}NDM & \cellcolor[HTML]{EFEFEF} & \cellcolor[HTML]{EFEFEF}\textbf{8.58}         & \cellcolor[HTML]{EFEFEF}\textbf{8.81}           & \cellcolor[HTML]{EFEFEF}48.41         & \cellcolor[HTML]{EFEFEF} & \cellcolor[HTML]{EFEFEF}\textbf{10.78}         & \cellcolor[HTML]{EFEFEF}\textbf{11.06}           & \cellcolor[HTML]{EFEFEF}62.12        \\
\bottomrule
\end{tabular}
\end{table}

Here we provide Table \ref{tab:results_ext} which contains additional resalts to Table \ref{tab:results}. This table compare DDPM and NDMs with learnable transformations on CIFAR-10 and ImageNet $32 \times 32$ datasets with different numbers of steps.

\subsection{Additional samples}

In this section, we present additional illustrations showcasing the properties of NDMs.

\begin{figure}[h]
    \centering
    \begin{subfigure}{\textwidth}
        \includegraphics[width=\textwidth]{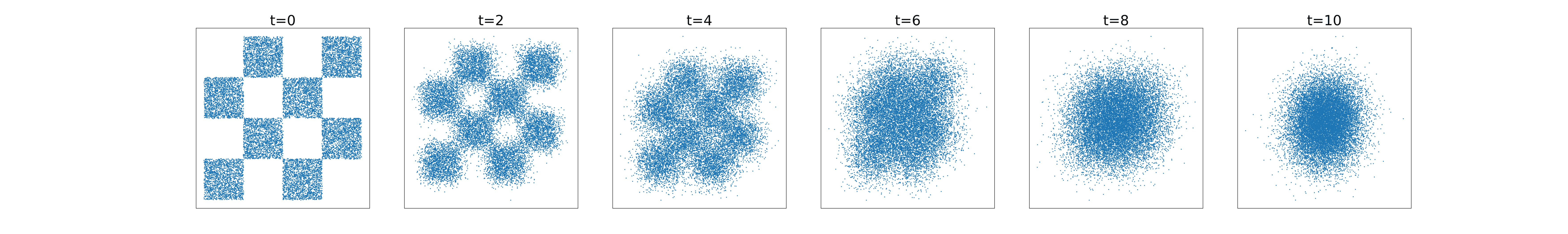}
        \caption{DDPM, samples $\z_t$ from forward process.}
        \label{fig:checker_comparison_ddpm_forward}
    \end{subfigure}\\
    \begin{subfigure}{\textwidth}
        \includegraphics[width=\textwidth]{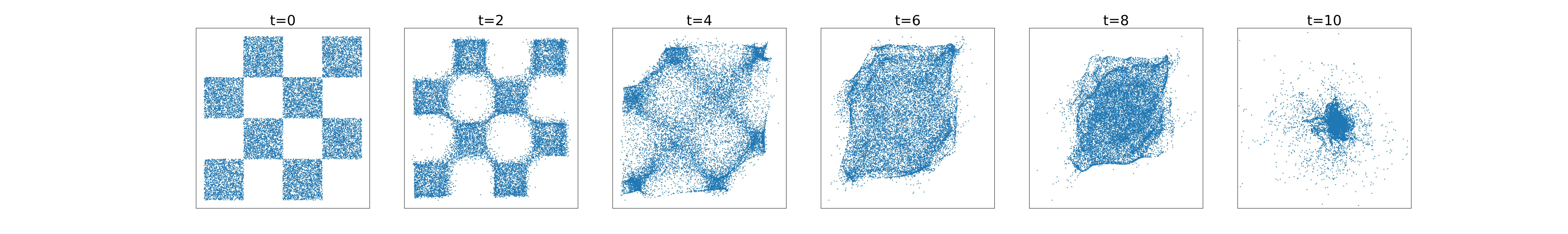}
        \caption{DDPM, predictions $\hx_{\theta}(\z_t, t)$ for different time steps.}
        \label{fig:checker_comparison_ddpm_predictions}
    \end{subfigure}\\
    \begin{subfigure}{\textwidth}
        \includegraphics[width=\textwidth]{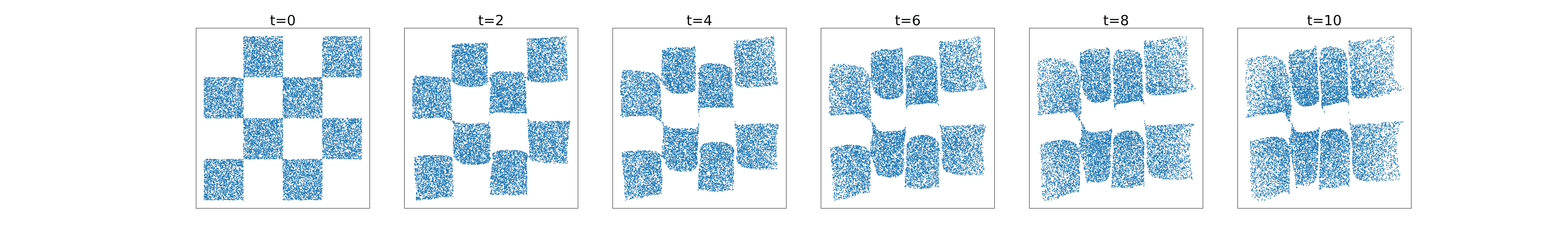}
        \caption{NDM, forward transformations $F_\vphi(\x, t)$.}
        \label{fig:checker_comparison_ndm_transform}
    \end{subfigure}\\
    \begin{subfigure}{\textwidth}
        \includegraphics[width=\textwidth]{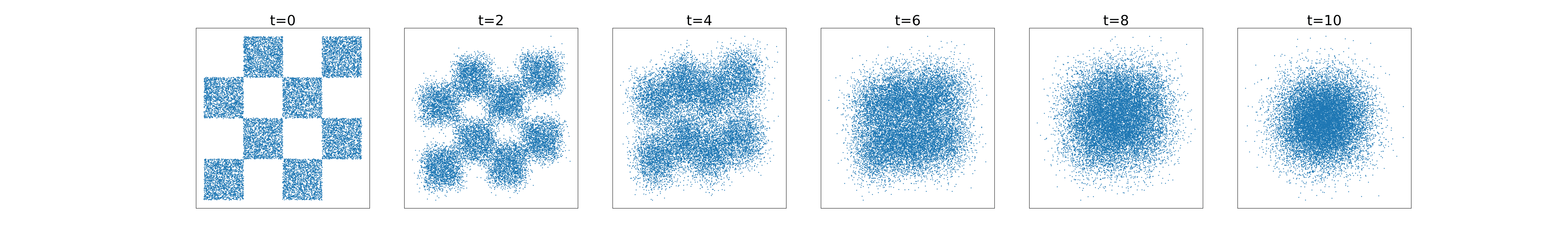}
        \caption{NDM, samples $\z_t$ from forward process.}
        \label{fig:checker_comparison_ndm_forward}
    \end{subfigure}
    \begin{subfigure}{\textwidth}
        \includegraphics[width=\textwidth]{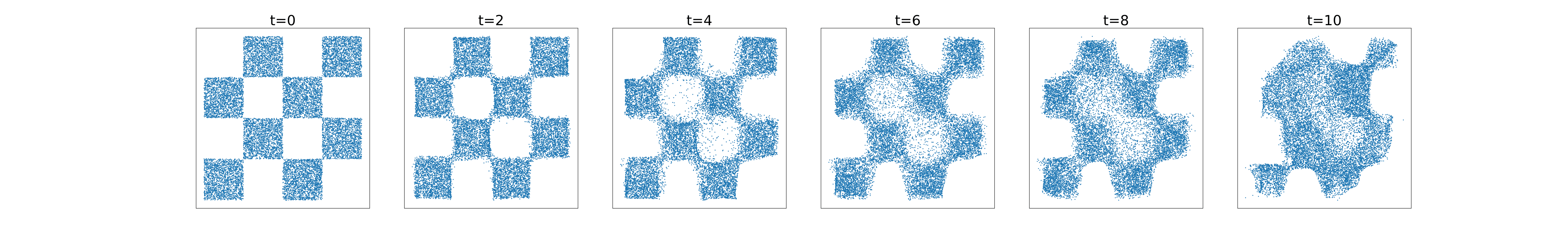}
        \caption{NDM, predictions $\hx_{\theta}(\z_t, t)$ for different time steps.}
        \label{fig:checker_comparison_ndm_predictions}
    \end{subfigure}
    \caption{Comparison of DDPM and NDM on 2D distribution.}
    \label{fig:checker_comparison}
\end{figure}

Figure \ref{fig:checker_comparison} provides a comparison between DDPM and NDM on a synthetic 2D data distribution. For this experiment, both models utilize $T=10$ discrete time steps. From Figure \ref{fig:checker_comparison_ndm_transform}, it is evident that NDM learns to transform the data distribution. Additionally, after injecting noise (Figures \ref{fig:checker_comparison_ddpm_forward} and \ref{fig:checker_comparison_ndm_forward}), the distributions of samples $\z_t$ show minimal differences between DDPM and NDM. However, when examining the predictions of data points $\hx_{\theta}(\z_t, t)$ (Figures \ref{fig:checker_comparison_ddpm_predictions} and \ref{fig:checker_comparison_ndm_predictions}), NDM produces predictions that more closely resemble the true data distribution compared to DDPM.

\begin{figure}[h]
    \centering
    \parbox{.495\textwidth}{
        \begin{subfigure}[b]{\linewidth}
            \includegraphics[width=\textwidth]{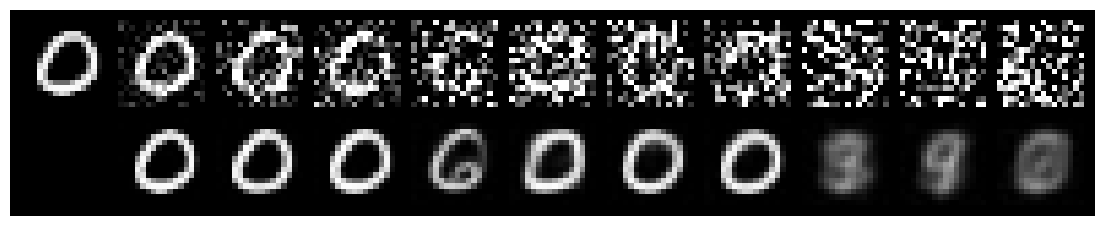}
        \end{subfigure}\\
        \begin{subfigure}[b]{\linewidth}
            \includegraphics[width=\textwidth]{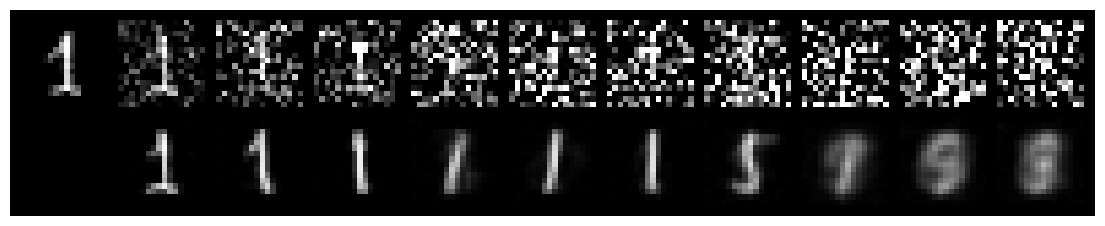}
        \end{subfigure}\\
        \begin{subfigure}[b]{\linewidth}
            \includegraphics[width=\textwidth]{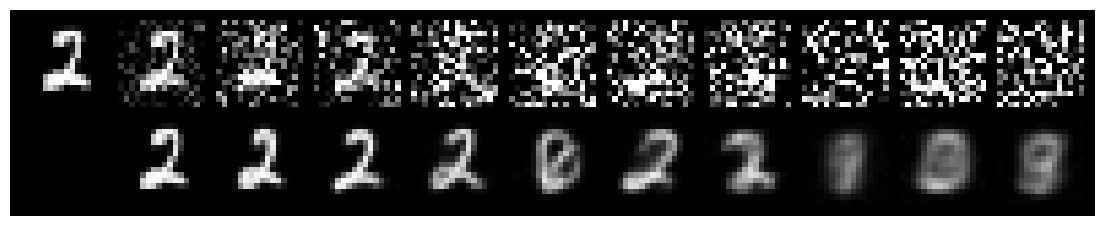}
        \end{subfigure}\\
        \begin{subfigure}[b]{\linewidth}
            \includegraphics[width=\textwidth]{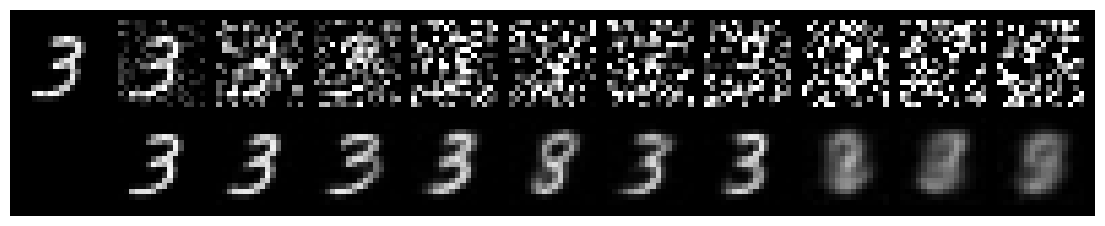}
        \end{subfigure}\\
        \begin{subfigure}[b]{\linewidth}
            \includegraphics[width=\textwidth]{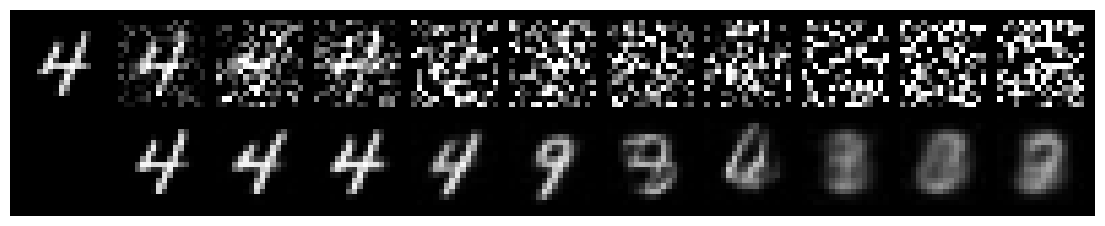}
        \end{subfigure}\\
        \begin{subfigure}[b]{\linewidth}
            \includegraphics[width=\textwidth]{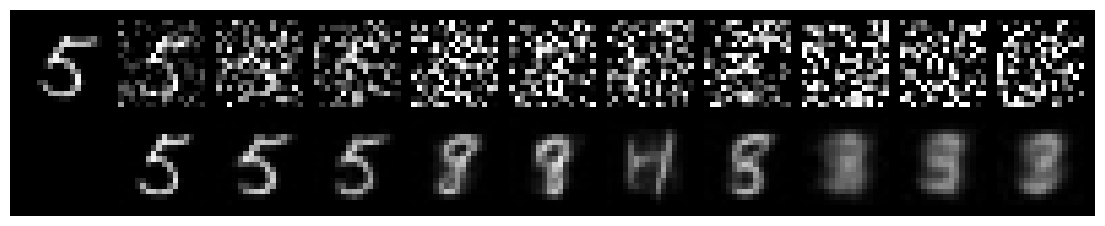}
        \end{subfigure}\\
        \begin{subfigure}[b]{\linewidth}
            \includegraphics[width=\textwidth]{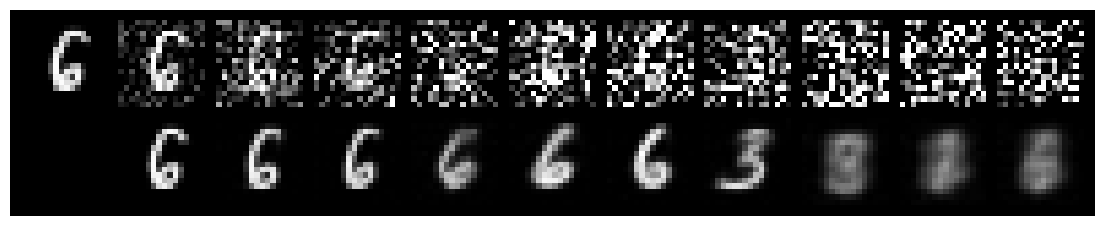}
        \end{subfigure}\\
        \begin{subfigure}[b]{\linewidth}
            \includegraphics[width=\textwidth]{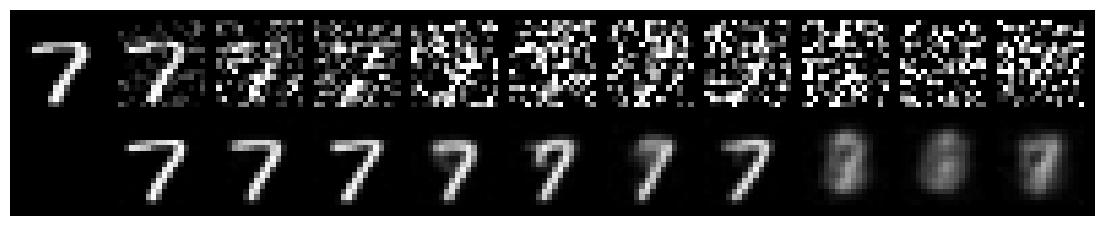}
        \end{subfigure}\\
        \begin{subfigure}[b]{\linewidth}
            \includegraphics[width=\textwidth]{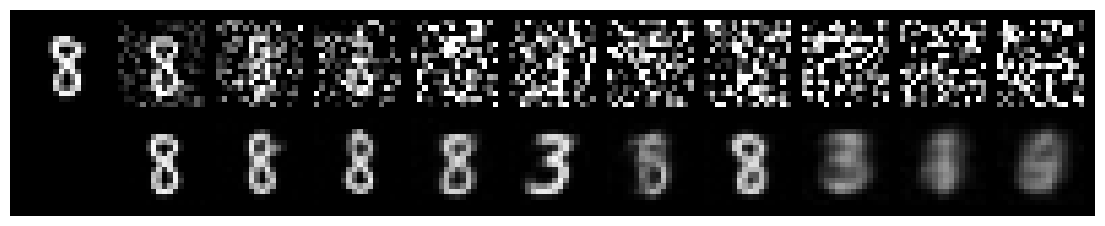}
        \end{subfigure}\\
        \begin{subfigure}[b]{\linewidth}
            \includegraphics[width=\textwidth]{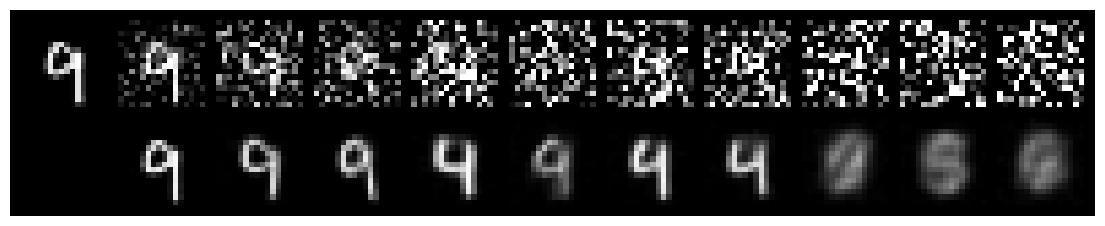}
            \caption{DDPM}
        \end{subfigure}
    }
    \hfill
    \parbox{.495\textwidth}{
        \begin{subfigure}[b]{\linewidth}
            \includegraphics[width=\textwidth]{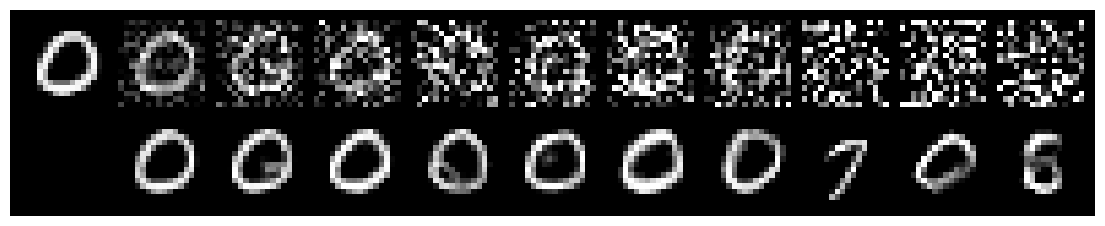}
        \end{subfigure}\\
        \begin{subfigure}[b]{\linewidth}
            \includegraphics[width=\textwidth]{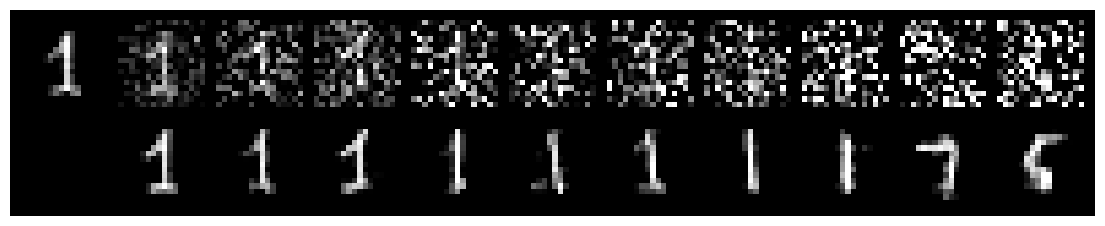}
        \end{subfigure}\\
        \begin{subfigure}[b]{\linewidth}
            \includegraphics[width=\textwidth]{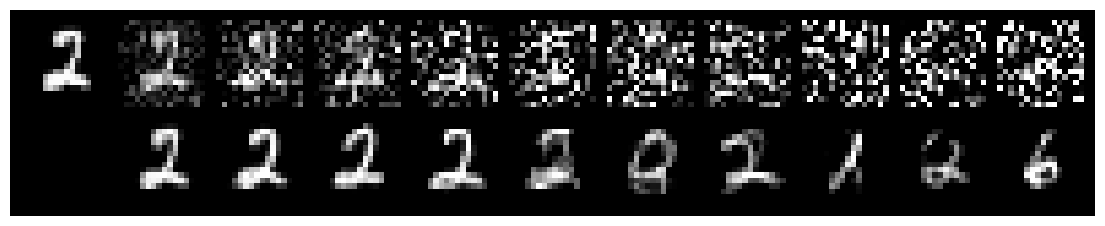}
        \end{subfigure}\\
        \begin{subfigure}[b]{\linewidth}
            \includegraphics[width=\textwidth]{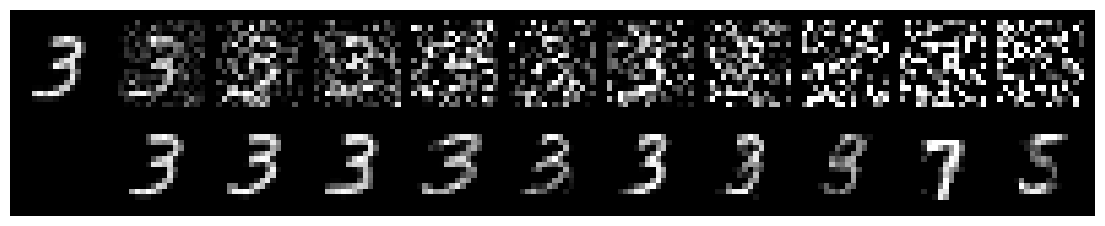}
        \end{subfigure}\\
        \begin{subfigure}[b]{\linewidth}
            \includegraphics[width=\textwidth]{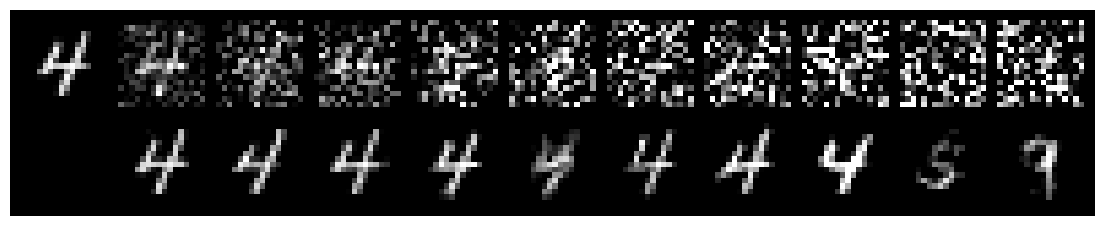}
        \end{subfigure}\\
        \begin{subfigure}[b]{\linewidth}
            \includegraphics[width=\textwidth]{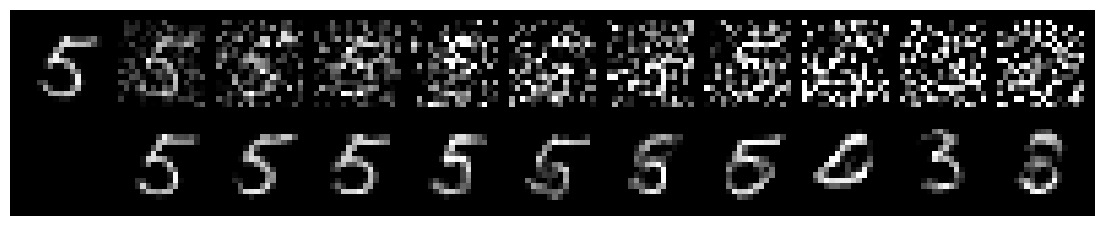}
        \end{subfigure}\\
        \begin{subfigure}[b]{\linewidth}
            \includegraphics[width=\textwidth]{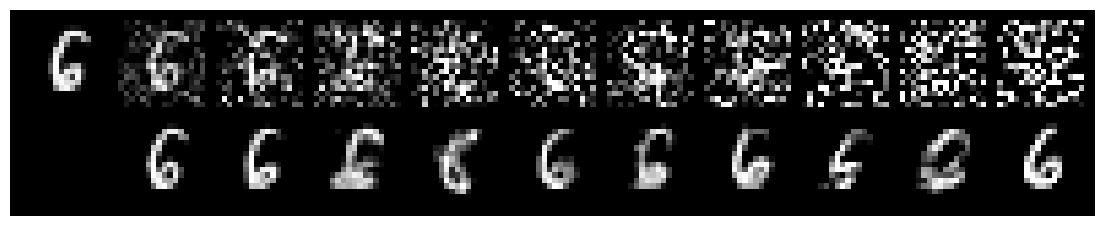}
        \end{subfigure}\\
        \begin{subfigure}[b]{\linewidth}
            \includegraphics[width=\textwidth]{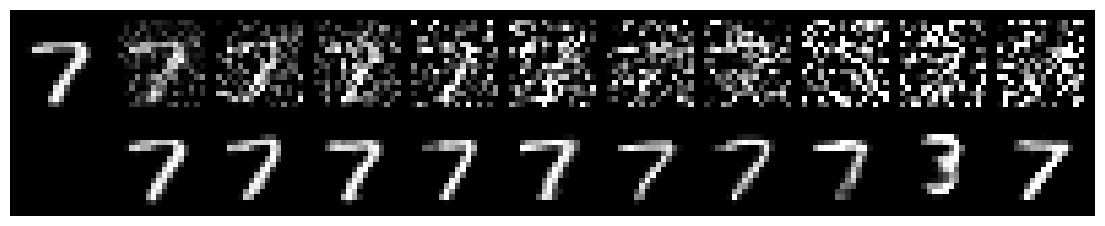}
        \end{subfigure}\\
        \begin{subfigure}[b]{\linewidth}
            \includegraphics[width=\textwidth]{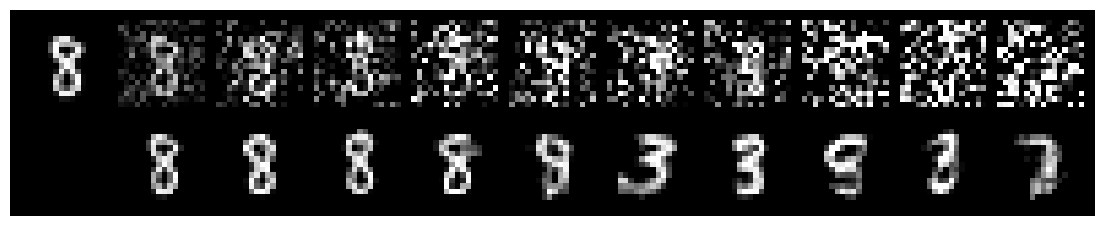}
        \end{subfigure}\\
        \begin{subfigure}[b]{\linewidth}
            \includegraphics[width=\textwidth]{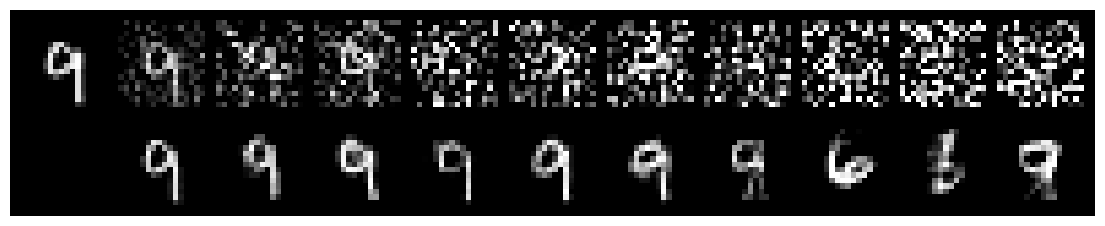}
            \caption{NDM}
        \end{subfigure}
    }
    \caption{Samples $\z_t$ from forward process and predicted data points $\hx_{\theta}(\z_t, t)$ on MNIST. (a) Samples from DDPM. (b) Samples from NDM. In each group, \emph{Left:} data sample, \emph{Top:} noised samples $\z_t$, \emph{Bottom:} predicted data points $\hx_{\theta}(\z_t, t)$.}
    \label{fig:predictions_mnist}
\end{figure}

A similar pattern emerges when applying these models to the MNIST dataset, as depicted in Figure \ref{fig:predictions_mnist}. For this experiment we also use $T=10$ discrete time steps. DDPM generates blurry predictions $\hx_{\theta}(\z_t, t)$ for $t$ close to $T$, which bear little resemblance to real MNIST samples. Conversely, NDM produces predictions that are more similar to the true MNIST distribution, despite both models generating similar-looking noisy samples.

\begin{figure}[h]
    \centering
    \begin{subfigure}{\textwidth}
        \includegraphics[width=\textwidth]{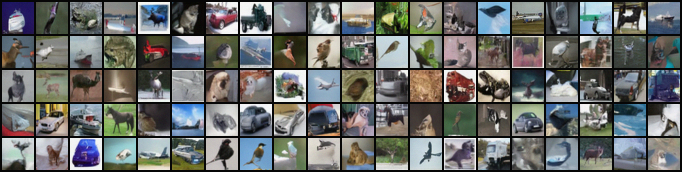}
        \caption{DDPM, FID = 11.44}
    \end{subfigure}\\
    \begin{subfigure}{\textwidth}
        \includegraphics[width=\textwidth]{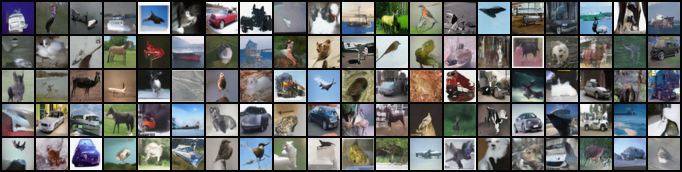}
        \caption{NDM, FID = 11.82}
    \end{subfigure}
    \caption{Samples on CIFAR-10. (a) Samples from DDPM. (b) Samples from NDM. Samples of both models are generated with the same random seed.}
    \label{fig:samples_cifar10}
\end{figure}

Finally, we include samples from both DDPM and NDM models with $T=1000$ steps on the CIFAR-10 dataset in Figure \ref{fig:samples_cifar10}. As outlined in Table 1, NDM exhibits lower sample quality based on FID measurements; however, visually there is no drop in quality.

\subsection{Ablation studies}

Finally, we address the question of whether the improved performance of NDM is due to the proposed method or merely the result of increasing the number of model parameters. To investigate this issue, we provide additional experiments where we double the number of DDPM parameters in two ways. The first way is to simply stack two U-Net architectures, which is the closest form to NDM. The second way is to increase the width of the U-Net architecture. Specifically, for the second way we use  $384$ channels instead of $256$. Importantly, we left all other hyper-parameters (see Table \ref{tab:hyper_parameters}), such as the learning rate and number of iterations, unchanged. As shown in Table \ref{tab:ablations}, neither of these approaches yields the same results as NDM with learnable transformations. This means that the improved performance is not simply a result of the increased number of parameters.

\section{Dynamic optimal transport}
\label{app:dot}

In this section, we present a proof-of-concept experiment demonstrating that the NDMs framework enables the learning of simpler generative trajectories. Specifically, we conduct experiments involving a 1D mixture of Gaussian distribution and dynamic optimal transport (OT).

While NDMs don't inherently have a direct connection with OT, we can establish a connection given the presence of infinitely many pairs of matched forward and reverse processes. This connection is facilitated by the NDMs' ability to learn the forward process. Therefore, we can consider the following setup.

We consider NDMs with a learnable function $F_\vphi$. Then, we constrain the reverse process to exclusively learn dynamic OT mappings. Finally, we train both the forward and reverse processes jointly, following the NDMs framework. In such a setup we can expect the forward process to learn such a transition from data distribution to Gaussian distribution, that aligns with the limitations imposed on the reverse process.

\subsection{Restricted reverse process}

\begin{table}[!t]
\caption{Comparison of NDM and DDPM with doubled number of parameters on CIFAR-10 for 10 and 1000 steps. The performance of DDPM stays the same while doubling the number of parameters, and NDM still achieves the best NLL and NELBO despite comparable number of parameters.}
\label{tab:ablations}
\centering
\begin{tabular}{llrrrrrrr}
\toprule
             &  & \multicolumn{3}{c}{\textbf{10 steps}}                                                                                & \multicolumn{1}{c}{\textbf{}} & \multicolumn{3}{c}{\textbf{1000 steps}}                                                                              \\ \cmidrule(r){3-5}  \cmidrule(r){7-9}
Model        &  & \multicolumn{1}{c}{NLL $\downarrow$} & \multicolumn{1}{c}{NELBO $\downarrow$} & \multicolumn{1}{c}{FID $\downarrow$} & \multicolumn{1}{c}{}          & \multicolumn{1}{c}{NLL $\downarrow$} & \multicolumn{1}{c}{NELBO $\downarrow$} & \multicolumn{1}{c}{FID $\downarrow$} \\ \cmidrule(r){1-1} \cmidrule(r){3-5}  \cmidrule(r){7-9}
DDPM         &  & 5.02                                 & 5.13                                   & 37.83                                 &                               & 3.11                                 & 3.18                                   & 11.44                                 \\
DDPM (stack) &  & 5.02                                 & 5.13                                   & 38.05                                 &                               & 3.10                                 & 3.18                                   & 11.42                                 \\
DDPM (wide)  &  & 5.01                                 & 5.11                                   & 37.88                                 &                               & 3.11                                 & 3.17                                   & \textbf{11.39}                                 \\
NDM          &  & \textbf{4.63}                                 & \textbf{4.74}                                   & \textbf{31.56}                                 &                               & \textbf{3.02}                                 & \textbf{3.03}                                   & 11.82                                \\
\bottomrule
\end{tabular}
\end{table}

To restrict the reverse process we parameterise the reverse deterministic process to have linear trajectories: 
\begin{align}
    \z_t = h_{\theta}(t, \veps) = (1-t) \hx_\theta(\veps) + t \veps,
\end{align}
where $\veps$ is a sample drawn from a unit Gaussian distribution. Since we are working with smooth 1D distributions, it is enough for $\hx_\theta$ to be monotonically increasing, so the trajectories $\z_t$ correspond to dynamic OT. Which means that for any parameters $\theta$ the reverse process describes dynamic OT between the standard Gaussian distribution and another distribution (not necessarily exactly the target data distribution). In practice, we parameterize $\hx_\theta$ using the neural network proposed by \citet{kingma2021variational} for the parameterization of the Signal-to-Noise Ratio (SNR) function.

Then, we can derive an ordinary differential equation (ODE) for the reverse process:
\begin{align}
    d \z_t = \underbrace{\veps - \hx_{\theta} (\veps)\Big\vert_{\veps=h_{\theta}^{-1} (t, \z_t)}}_{f_{\theta} (t, \z_t)} d t.
\end{align}

Next, we may switch to a stochastic differential equation (SDE) according to \citet{song2021scorebased}:
\begin{align}
    d \z_t = \underbrace{\left[ f_{\theta} (t, \z_t) - \frac{g^2(t)}{2} \nabla_{\z_t} \log p_{\theta} (\z_t) \right]}_{f_{\theta} ^r (t, \z_t)} d t + g(t) d \bw.
    \label{eq:ot_rsde}
\end{align}

As soon as we have access to $h_{\theta}^{-1}$, we may find:
\begin{align}
    \nabla_{\z_t} \log p_{\theta} (\z_t) 
    &= \nabla_{\z_t} \left[ \log p(\veps) - \log \left| \frac{\d \z_t}{\d \veps} \right| \right] \Big\vert_{\veps=h_{\theta}^{-1} (t, \z_t)} \\
    &= \nabla_{\z_t} \left[ \log p(\veps) - \log \left| (1 - t) \frac{\d \x_t}{\d \veps} + t \right| \right] \Big\vert_{\veps=h_{\theta}^{-1} (t, \z_t)}.
\end{align}

\begin{figure}[tp]
    \centering
    \parbox{.49\textwidth}{
        \begin{subfigure}{\linewidth}
            \includegraphics[width=\textwidth]{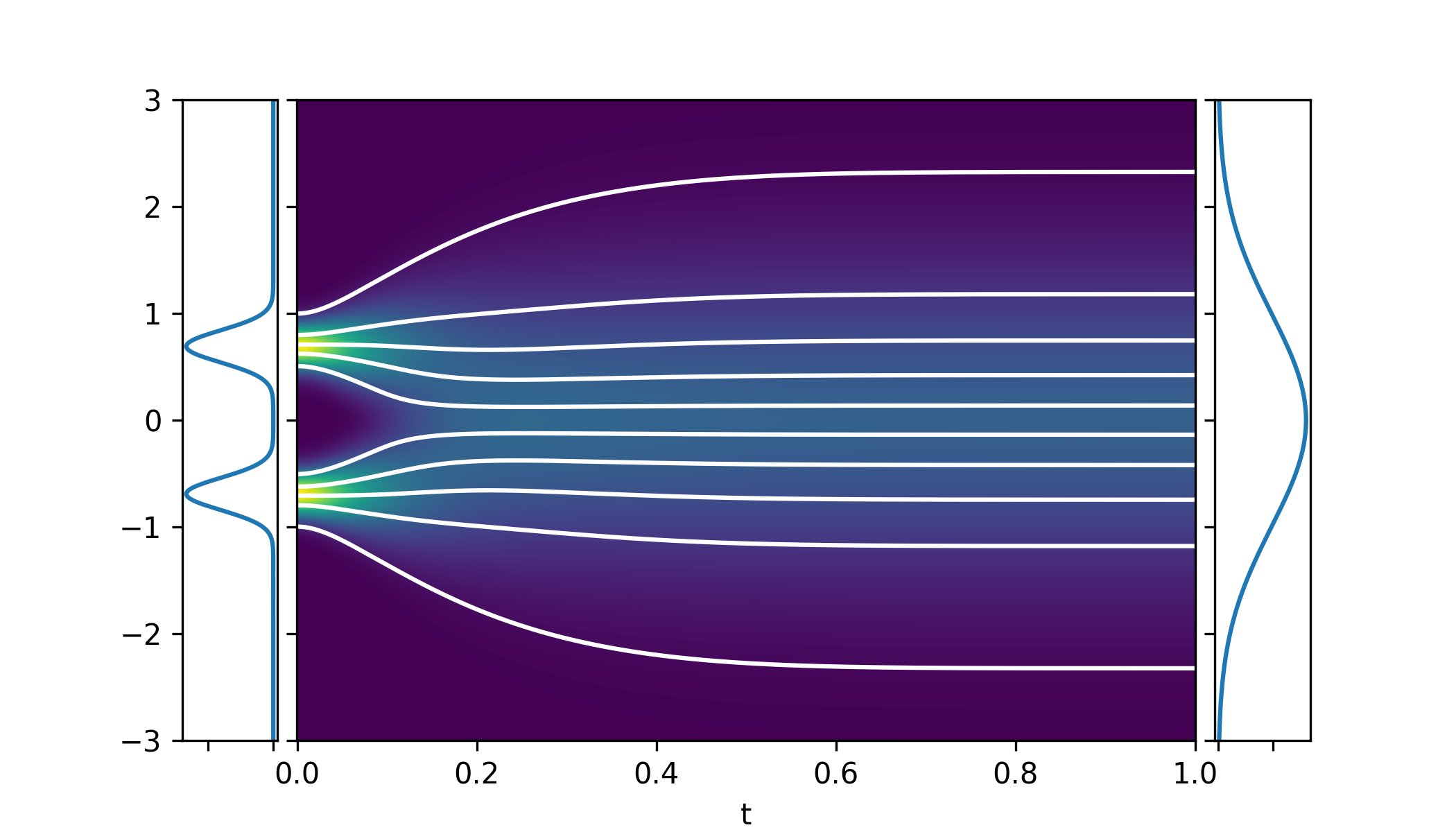}
            \caption{DDPM with regular reverse process.}
            \label{fig:dot_ddpm}
        \end{subfigure}
    }
    \parbox{.49\textwidth}{
        \begin{subfigure}{\linewidth}
            \includegraphics[width=\textwidth]{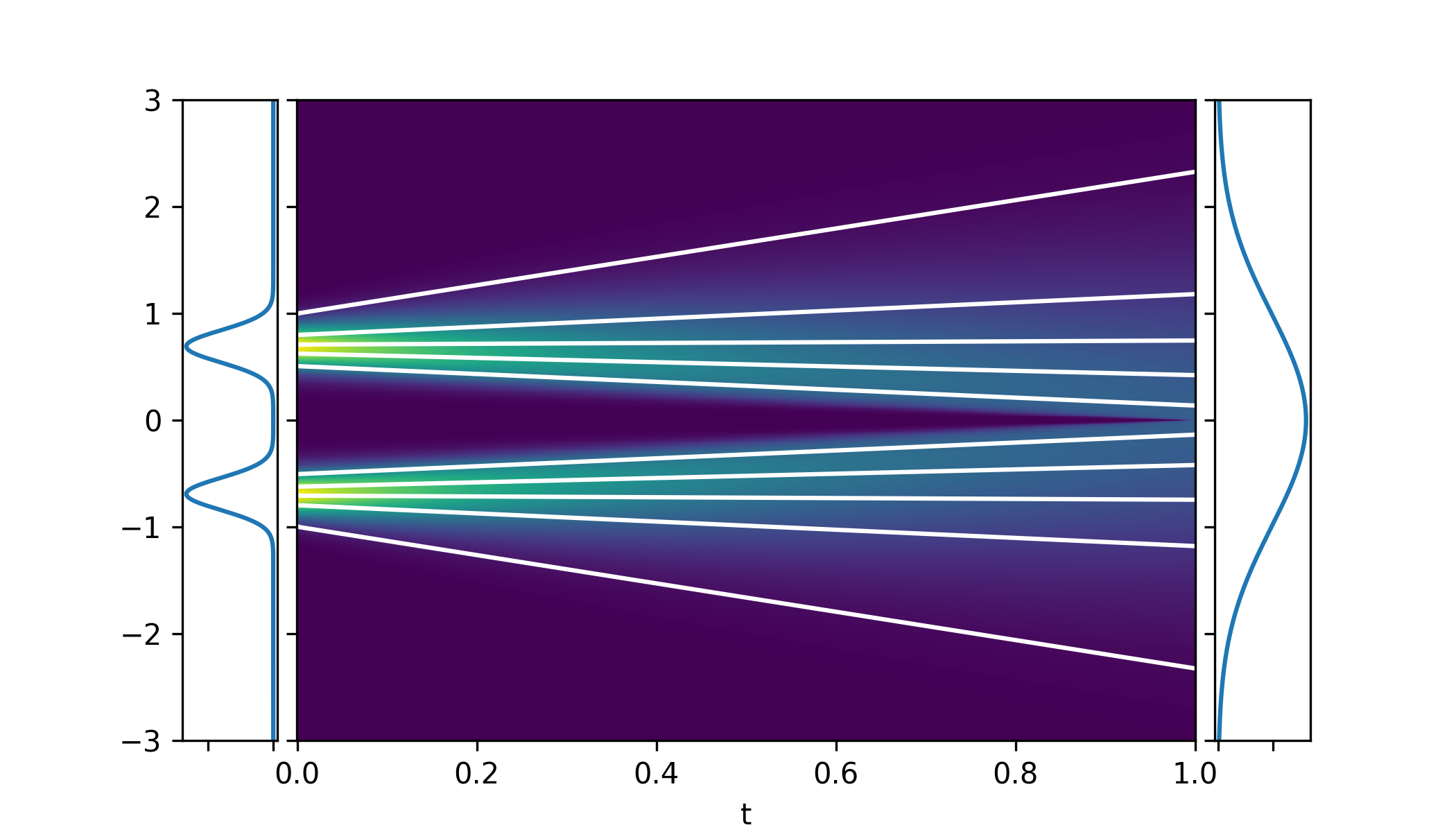}
            \caption{NDM with restricted (OT) reverse process.}
            \label{fig:dot_ndm}
        \end{subfigure}
    }
    \caption{Comparison of DDPM and NDM with restricted reverse process to be optimal transport, 1D distribution.}
\end{figure}

\subsection{Objective function}

To train a model with such a specific reverse process, we can utilize a slightly modified NDMs framework. The only component of the NDMs' objective that is unclear is the diffusion term $\Ldiff$. NDMs provide a conditional reverse SDE associated with the forward process (\ref{eq:forward_sde}) in the following form:
\begin{align}
    d \z_t = f^f_{\vphi} (x, t, \z_t) d t + g(t) d \bw.
\end{align}

Also, here we have the reverse SDE (\ref{eq:ot_rsde}). Therefore, we may find diffusion term $\Ldiff$ of objective as follows:
\begin{align}
    \Ldiff = \E_{q(\x)} \E_{u(t)} \E_{q(\z_t|\x)}
    \frac{1}{g^2(t)} \Bigg\| f^f_{\vphi} (x, t, \z_t) - f_{\theta} ^r (t, \z_t) \Bigg\|_2^2.
\end{align}

\subsection{Results and discussion}

Figures \ref{fig:dot_ddpm} and \ref{fig:dot_ndm} illustrate trajectories learned by DDPM and NDM with learnable $F_\vphi$ and restricted reverse process. As expected, DDPM learns curved trajectories predetermined by fixed forward process. At the same time NDM effectively learns dynamic OT. It worths noting that DDPM with the restricted reverse process is by design not able to learn the data distribution, since it’s impossible to match the fixed forward process (with curved trajectories) with the reverse process (with straight trajectories).

The proposed approach is limited to 1D data, monotonically increasing $\hx_\theta$, and a nontrivial $h_{\theta}^{-1}$ function, which we resolve using 5 iterations of Newton's method. Nevertheless, this experiment clearly demonstrates that NDMs may be utilised for learning OT as well as other (e.g. computationally efficient ones) dynamics by restricting the reverse process. Establishing rigorous theoretical connections with OT, developing specific techniques for efficient parameterisation of the reverse process and generalising to higher dimensions are interesting avenue for future work.

\newpage

\end{document}